\documentclass[10pt,twocolumn,letterpaper]{article}

 \usepackage[pagenumbers]{cvpr} 
\usepackage{amsmath,amsfonts}
\usepackage[algo2e,lined,boxed,commentsnumbered]{algorithm2e}
\usepackage{algorithmic}
 \usepackage{algorithm}

\usepackage{textcomp}
\usepackage{stfloats}
\usepackage{url}
\usepackage{subfloat}
\usepackage{verbatim}
\usepackage{graphicx}
\usepackage{float}
\usepackage{caption}
\usepackage{tabularx}
\usepackage{array}
\usepackage{booktabs}
\usepackage{multirow}
\usepackage{makecell}
\usepackage{tikz}
\usepackage[latin1]{inputenc}
\usepackage{pgfplots}
\pgfplotsset{compat=1.3}
\usetikzlibrary{backgrounds,scopes}   
\usetikzlibrary{spy}
\usetikzlibrary {arrows.meta} 
\usepackage{bm}
\usepackage{colortbl}

\definecolor{cvprblue}{rgb}{0.21,0.49,0.74}
\usepackage[pagebackref,breaklinks,colorlinks,citecolor=cvprblue]{hyperref}

\def\paperID{3043} 
\def\confName{CVPR}
\def\confYear{2025}

 \title{SmileSplat: Generalizable Gaussian Splats for Unconstrained Sparse Images}

\author{Yanyan Li$^{1}$ 
\and
Yixin Fang$^{2}$ 
\and
Federico Tombari$^{3,4}$
\and Gim Hee Lee $^{1}$
\and
$^{1}$National University of Singapore\\
$^{2}$Zhejiang university\\
$^{3}$Technical University of Munich \\
$^{4}$Google
}

\begin{document}
\maketitle
\begin{abstract}
\underline{S}parse \underline{M}ulti-view \underline{I}mages can be \underline{Le}arned to predict explicit radiance fields via Generalizable Gaussian \underline{Splat}ting approaches, which \textcolor{black}{
can achieve} wider application prospects in real-life when ground-truth camera parameters are not required as inputs.
In this paper, a novel generalizable Gaussian Splatting method, \textbf{SmileSplat}, is proposed to reconstruct pixel-aligned Gaussian surfels for diverse scenarios only requiring unconstrained sparse multi-view images.
First, Gaussian surfels are predicted based on the multi-head Gaussian regression decoder, which can are represented with less degree-of-freedom but have better multi-view consistency. Furthermore, the normal vectors of Gaussian surfel are enhanced based on high-quality of normal priors. 
Second, the Gaussians and camera parameters (both extrinsic and intrinsic) are optimized to obtain high-quality Gaussian radiance fields for novel view synthesis tasks based on the proposed Bundle-Adjusting Gaussian Splatting module.
Extensive experiments on novel view rendering and depth map prediction tasks are conducted on public datasets, demonstrating that the proposed method achieves state-of-the-art performance in various 3D vision tasks. More information can be found on our project page (\url{https://yanyan-li.github.io/project/gs/smilesplat}).
\end{abstract}    
\section{Introduction}~\label{sec:intro}


Novel view rendering technology has become widely used in both industrial production and entertainment applications, enabling the generation of photo-realistic views of unknown environments based on estimated implicit or explicit radiance fields~\cite{mildenhall2021nerf, kerbl20233d}. To construct high-quality radiance fields, Structure-from-Motion (SfM)\cite{schonberger2016structure,pan2024global} or Simultaneous Localization and Mapping (SLAM)~\cite{li2021rgb,yuan2022orb} systems are typically employed to generate 3D point clouds, along with camera intrinsics and extrinsics, from a set of dense-view RGB images. However, when working with sparse-view images ( as few as two), accurately estimating 3D point primitives and camera parameters using traditional systems becomes challenging. This issue is particularly pronounced in low-textured scenes or with fast camera motion, where the limited number of images fails to provide sufficient constraints for conventional optimization techniques.

\begin{figure}
    \centering
    \includegraphics[width=\linewidth]{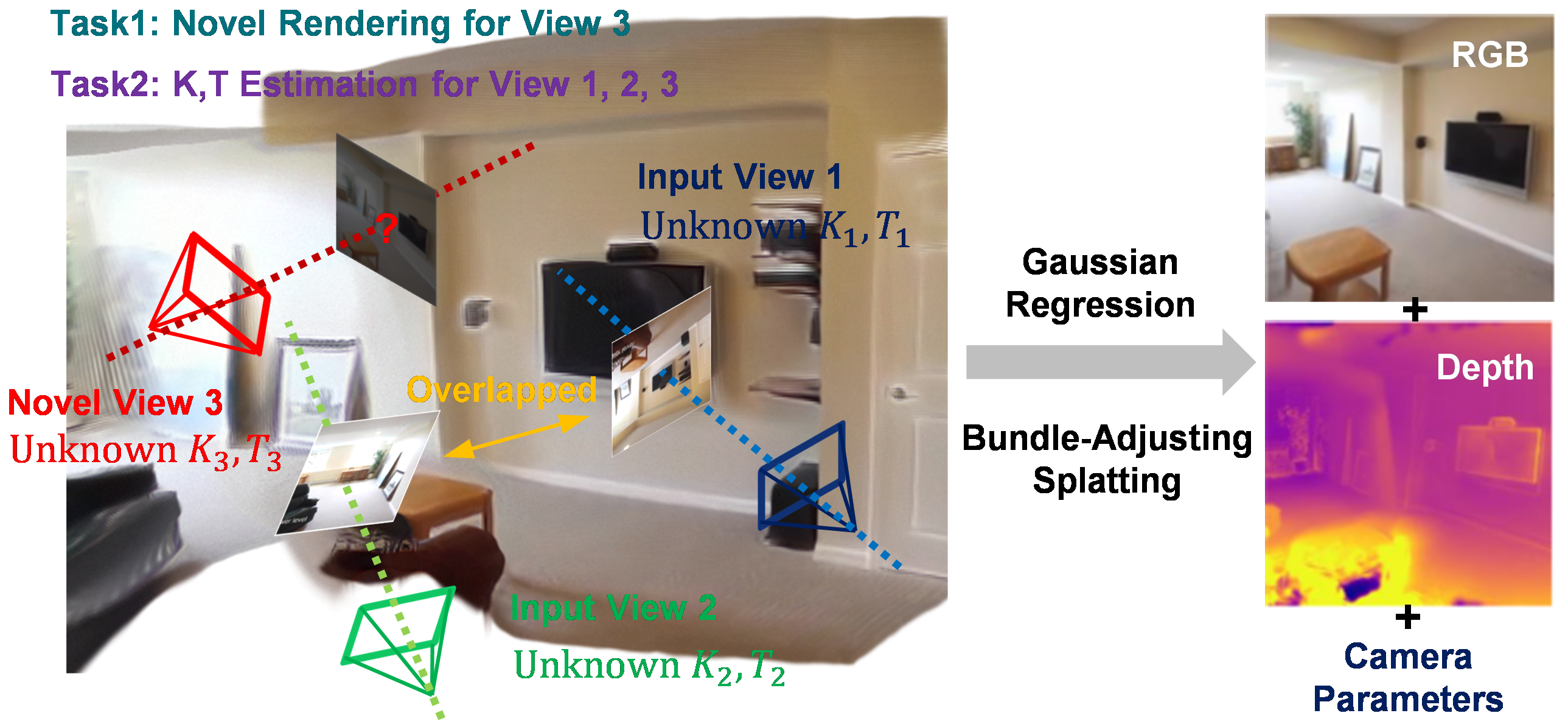}
    \caption{ An example result of SmileSplat. It aims to render novel views and estimate camera parameters (intrinsic \(\mathbf{K}\) and extrinsic \(\mathbf{T}\)) by using sparse views.}
    \label{fig:enter-label}
\end{figure}

Camera calibration~\cite{zhang2000flexible}, tracking~\cite{mur2015orb}, dense mapping~\cite{hirschmuller2005accurate, kazhdan2006poisson}, and rendering~\cite{oechsle2019texture} are classic problems in the fields of computer vision and computer graphics, typically addressed by traditional solutions based on multi-view geometry strategies~\cite{mur2015orb,schonberger2016structure,kazhdan2006poisson},  
Compared to traditional methods, Neural Radiance Fields (NeRFs)~\cite{mildenhall2021nerf} and 3D Gaussian Splatting~\cite{kerbl20233d} have achieved photo-realistic novel view rendering performance. Given 2D posed images, NeRFs~\cite{mildenhall2021nerf} address the problem by training neural networks to encode implicit 3D representations, enabling significant advances in photo-realistic novel view rendering applications. To improve efficiency in training and inference, 3D Gaussian Splatting methods use point-based representations instead of implicit neural parameterizations via differential optimization modules. 
Even though Gaussian Splatting SLAM approaches~\cite{matsuki2023gaussian,yan2023gs,zhu2024robust} have capabilities in tracking cameras in unknown scenes, they have limitations in two main areas. On the one hand, a large number of views are required for training. On the other hand, camera intrinsic parameters and depth information are necessary to achieve robust rendering performance when dealing with monocular RGB inputs. 
When the input images collected in challenging scenes are sparse, it is extremely difficult to calibrate camera parameters and obtain the 3D structure of the unknown scenes. Therefore, the traditional Gaussian Splatting~\cite{kerbl20233d,li2025geogaussian} and neural radiance fields~\cite{mildenhall2021nerf,barron2022mip} are doubly challenged by low-quality inputs and fewer constraints. 

To improve the performance of novel view rendering in sparse-shot tasks, generalizable Gaussian splatting approaches~\cite{yu2021pixelnerf,chen2024mvsplat,ye2024no} are explored in this area. These approaches can be classified into two types: \textbf{CamPara-Required} and \textbf{CamPara-Free} methods, depending on whether camera parameters are required as part of the inputs.
For CamPara-Required methods, given ground truth camera poses and intrinsics, 3D Gaussians predicted by networks~\cite{chen2024mvsplat, charatan2023pixelsplat} are rendered to novel viewpoints. To generate accurate Gaussian primitives, neural multi-head decoders first predict depth (point clouds), covariance, and opacity values. A forward-stream rendering module is then utilized to optimize the initial Gaussian parameters based on these predictions.
To further eliminate the reliance on camera parameters, point clouds, rather than depth maps, are predicted by networks such as Dust3R~\cite{wang2023dust3r} and Mast3R~\cite{mast3r_arxiv24} in the canonical system. These point clouds are used to initialize 3D Gaussians~\cite{fan2024instantsplat}, which are subsequently to obtain camera poses. Then, Gaussian parameters will be optimized through the Gaussian Splatting module~\cite{kerbl20233d}. Since these two modules are separate, additional iterations are required to train the 3D Gaussians for such scenarios.
Compared to CamPara-Required methods, CamPara-Free ones are more convenient for applications since they do not require an initialization step. However, these pioneering CamPara-Free approaches tend to directly integrate camera calibration based on predicted point clouds and the Gaussian Splatting module together, which limits the further high-quality achievement of generalizable Gaussian radiance fields.

\begin{figure*}
    \centering
    \includegraphics[width=\textwidth]{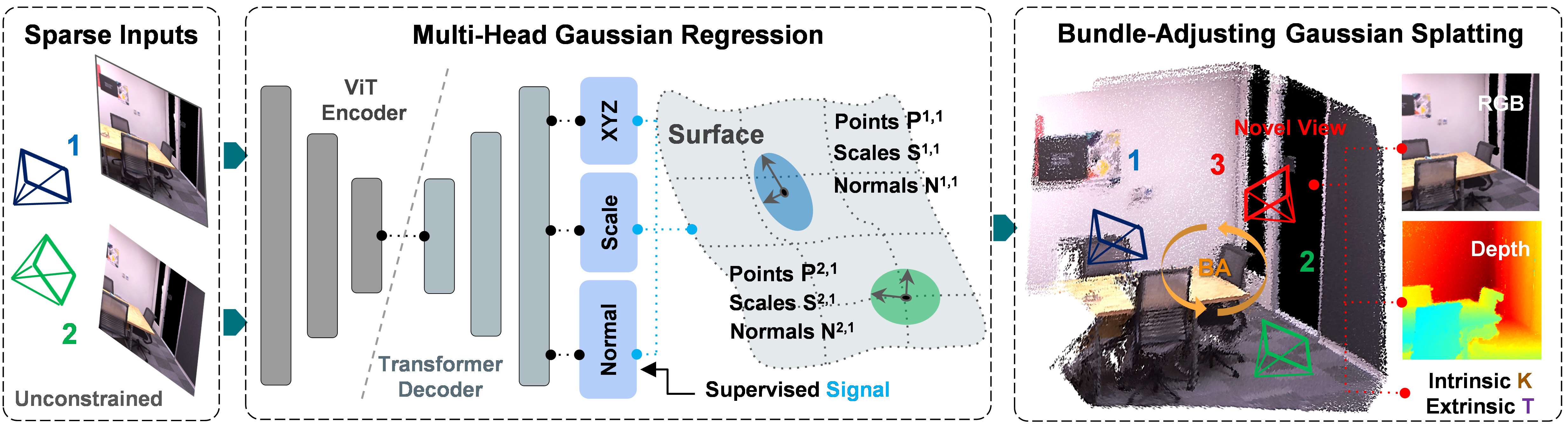}
    \caption{Architecture of SmileSplat. With sparse but overlapping views as input, the system consists of two main modules, Multi-Head Gaussian Regression and  Bundle-Adjusting Gaussian Splatting, for achieving scaled Gaussian radiance fields.}
    \label{fig:archi}
\end{figure*}

In this paper, we propose a new generalizable Gaussian Splatting architecture, as shown in Figure~\ref{fig:archi}, aimed at achieving high-fidelity novel view rendering performance for unconstrained sparse-view images.
First, Gaussian surfels are predicted by a forward-stream neural network that utilizes a standard transformer encoder (Siamese ViT encoders~\cite{dosovitskiy2020image} and cross-attention embedding blocks~\cite{weinzaepfel2022croco}) to detect geometric priors from images. These deep priors, along with the images, are then fed into the proposed Multi-head Gaussian Regression Decoder to predict pixel-aligned, generalizable 3D Gaussian surfel parameters in the canonical coordinate frame.
Furthermore, we estimate an intrinsic matrix by considering geometric and photometric Constraints based on the initial 3D Gaussians. To the best of our knowledge, our method is the first to render images without requiring predefined intrinsic parameters. Next, the relative extrinsic matrix between images is predicted based on the estimated Gaussian surfels.
To improve the consistency of the predicted Gaussian surfels, we propose a Gaussian Splatting Bundle Adjustment method to further refine the Gaussian parameters, intrinsics, and extrinsics. This refinement is based on photometric and geometric constraints, allowing us to establish scaled Gaussian Radiance Fields for unconstrained sparse images.
The contributions of this paper are summarized as:
\begin{enumerate}
    \item We present a versatile generalizable Gaussian Splatting architecture for un-calibrated and un-posed sparse-view images.
    \item A camera parameter optimization module based on Gaussian Splatting is analyzed to achieve accurate motion estimation of sparse images. 
    \item A bundle-adjusting Gaussian Splatting algorithm is implemented  to produce high-quality and scaled Gaussian radiance fields. 
    \item Extensive evaluations on variety of scenes demonstrate competitive novel view rendering performance. The code and generated dataset will be released to the community.
\end{enumerate}
\section{Related Work}~\label{sec:related_work}

\paragraph{Radiance Fields for a Single Scene.} 
Radiance Fields have garnered significant attention in the field of 3D vision due to their ability to generate novel views of objects or scenes from arbitrary viewpoints. Neural Radiance Fields (NeRF)~\cite{mildenhall2021nerf} is among the pioneering and most prominent methods for efficiently rendering high-quality novel views by representing 3D scenes implicitly via Multi-layer Perceptrons (MLPs). However, NeRF is hindered by slow training and inference speeds. Subsequent research~\cite{barron2022mip, barron2023zip} has primarily focused on either enhancing rendering quality or improving computational efficiency. Recent advances have introduced explicit volume structures such as multi-resolution voxel grids~\cite{fridovich2022plenoxels,sun2022direct} or hash functions~\cite{muller2022instant} to improve performance. Despite these improvements, per-pixel ray marching remains a bottleneck for rendering speed, while which is a critical issue in SLAM applications requiring real-time map interactions. In contrast, 3D Gaussian Splatting (3DGS)~\cite{kerbl20233d} employs anisotropic 3D Gaussians to represent radiance fields, combined with differentiable splatting for rendering. This method excels at quickly reconstructing complex real-world scenes with high detail, particularly capturing high-frequency details effectively. By iterating over rasterized primitives instead of marching along rays, 3DGS utilizes the natural sparsity of 3D scenes, offering a balance of high-fidelity representation and efficient rendering. Various studies have applied 3D Gaussians and differentiable rendering to static scene capture~\cite{yan2023gs,cheng2024gaussianpro}, with recent works demonstrating superior results in dynamic scene capture~\cite{wu20234d,yang2023real}.

\paragraph{Generalizable Gaussian Splatting.}
Similar to multi-view stereo tasks, when two images with limited overlaps are fed into 3D Gaussian Splatting systems, several new challenges arise since traditional Gaussian Splatting systems~\cite{kerbl20233d} require a large number of images as inputs. SparseGS~\cite{xiong2023sparsegs} addresses these challenges by using diffusion networks to remove outliers and different types of rendered depth maps to detect floaters. Instead of relying on initial point clouds generated from COLMAP~\cite{fu2023colmap} or SLAM systems~\cite{li2021rgb}, MVSplat~\cite{chen2024mvsplat} and COLMAP-free GS~\cite{fu2023colmap} estimate depth maps for both target and source RGB images, where these depth values are then transferred to point clouds, and the mean vectors of Gaussian ellipsoids are initialized based on these point clouds. 
Benefiting from the point clouds of Dust3R~\cite{wang2023dust3r}, the Gaussian Splatting process in InstantSplat~\cite{fan2024instantsplat} becomes more efficient compared to randomly generated point clouds used in MonoGS~\cite{matsuki2023gaussian}. Based on the camera parameters estimated in Dust3r, InstantSplat optimizes 3D Gaussians and camera poses in the rendering process. However, the separation of modules means that there is a missed opportunity to further accelerate the convergence of Gaussian Splatting. 
Compared to InstantSplat which estimates camera parameters based on predicted point clouds from Dust3R, pixelSplat~\cite{charatan2023pixelsplat} predicts 3D Gaussians directly, which are defined in the Gaussian Splatting rasterization using ground truth camera parameters. Recently, pose-free novel view rendering methods~\cite{ye2024no} have estimated and optimized camera poses using the PnP algorithm~\cite{hartley2003multiple} and Gaussian Splatting SLAM method~\cite{matsuki2023gaussian}, while incorporating intrinsic parameters into a deep feature token to predict scenes at a reasonable scale.
Unlike these approaches, our method predicts 3D Gaussians for uncalibrated and unposed images, estimating the unknown camera parameters based on the predicted Gaussians. All these estimated parameters and representations are then optimized via the bundle-adjusting Gaussian Splatting module to establish high-fidelity radiance fields.

\section{Methodology}~\label{sec:method}

\subsection{Gaussian Surfel Prediction}

Instead of using 3D Gaussians to represent 3D scenes, as used in PixelSplat~\cite{charatan2023pixelsplat} and MVSplat~\cite{chen2024mvsplat}, the network in this paper predicts Gaussian surfels~\cite{yan2023gs} for two main reasons. First, 3D Gaussians are not ideal for sparse-view training tasks due to limited constraints on Gaussian consistency~\cite{huang20242d}. Second, Gaussian surfels involve fewer parameters compared to the rotation matrix, and the more important thing is that they are easier for training compared to the covariance matrix, as several robust large language models~\cite{bae2024dsine} can be leveraged to train this head.



\paragraph{Gaussian Surfel Representation.}

Each 3D Gaussian surfel \(\mathcal{G}_i\) is parameterized by 12 values, consisting of three for its RGB color \(\mathbf{c}_i\), three for its center position \(\bm{\mu}_i = [x_i \; y_i \; z_i]^T\), two for its scale vector \(\mathbf{s}_i = [s^x_i \; s^y_i]^T\), three for its normal vector \(\mathbf{n}_i = [n^x_i \; n^y_i \; n^z_i]^T\) representing its direction, and one for its opacity \(o_i \in [0, 1]\). These parameters are expressed as:
\begin{equation}
\mathcal{G}_i = [\mathbf{c}_i\; \bm{\mu}_i \; \mathbf{s}_i \;\mathbf{n}_i \; o_i ].
\end{equation}

The covariance matrix \(\bm{\Sigma}_i\) for the Gaussian surfel can then be defined as:
\begin{equation}
    \bm{\Sigma}_i = \mathbf{R}_i Diag[(s^x_i)^2 \; (s^y_i)^2 \; 0] (\mathbf{R}_i)^T
    \label{eq_sigma}
\end{equation}
where \(\mathbf{R}_i = [\mathbf{n}_1 \; \mathbf{n}_2 \; \mathbf{n}_i] \in SO(3) \) is an orthonormal matrix. Here, \(\mathbf{n}_1\) lies on the great circle of \(\mathbf{n}_i\), and \(\mathbf{n}_2\) is obtained via the cross product \(\mathbf{n}_1 \times \mathbf{n}_i\). $ Diag[\cdot]$ is a diagonal matrix.


\paragraph{Standard Transformer Backbone.}
Inspired by the architecture of CroCo~\cite{weinzaepfel2022croco} and Dust3D~\cite{wang2023dust3r}, we use the model that is pre-trained based on million-level images, where two ViT~\cite{dosovitskiy2020image} encoders sharing weights are fed by multi-view stereo images $I_1$ and $I_2$. Continually, a generic transformer network~\cite{weinzaepfel2023croco} equipped with self-attention and cross-attention strategies is used to deal with tokens from the Siamese ViT backbone.

\paragraph{Multi-head Gaussian Regression Decoder.}

In the Gaussian regression head, we predict 3D Gaussians in the coordinate system of image \(I_1\), using four separate blocks to estimate the position \(\mathbf{P}\), surface normal \(\mathbf{n}\), opacity \(o\), and scaling \(\mathbf{s}\) vectors. To reduce the parameter scale, we directly use the color of each pixel in the Gaussian surfel representation. 

For the position prediction, we use an offset prediction block, leveraging a pretrained pointmap head from Dust3R~\cite{wang2023dust3r}. The position is expressed as \(\mathbf{P} = \hat{\mathbf{P}} + \delta \mathbf{P}\), where \(\hat{\mathbf{P}}\) is the predicted position and \(\delta \mathbf{P}\) is the offset, predicted by a 3-layer MLP network fed with the concatenation of pointmap features and the encoder embeddings.

For the surface normal prediction, we employ a U-Net architecture that predicts surface normals for each pixel based on the input image and deep embeddings. 
To achieve faster convergence of this block, we use the predicted surface normals \(\hat{\mathbf{n}}\) from a pre-trained surface normal network~\cite{bae2024dsine} to supervise the normal prediction head. The supervision is applied using the following loss function:
\begin{equation}
   \mathcal{L}_{\mathbf{n}} = | \mathbf{n} - \hat{\mathbf{n}} |_{1} + |1- \mathbf{n}\cdot \hat{\mathbf{n}}|_{1}.
\end{equation}

To predict opacity, we note that points with low offsets are more likely to be on the accurate surface and should thus have higher opacity. Therefore, we feed the deep features from the backbone and features from the position offset block to predict the opacity of each 3D Gaussian.

For the shape prediction of each Gaussian, we already have its surface normal, then we estimate the scaling component of the Gaussian. Benefitting of the using representation, the scale vector is defined as $\mathbf{s}= [s_1 \; s_2 \; 0]$, which is predicted by a 2-layer MLP architecture with the inputs of pointmaps and deep embeddings. Therefore, the covariance matrix of the Gaussian surfel can be represented based on the Equation~\ref{eq_sigma}.


\subsection{Camera Parameter Optimization Based on a Single View}
\label{sec:camera_para_op}
\paragraph{Intrinsic Parameter Estimation.} 
In methods like COLMAP~\cite{schoenberger2016mvs, schoenberger2016sfm} and Dust3R~\cite{wang2023dust3r}, intrinsic parameters are typically estimated by solving optimization problems that relate pixels to point clouds in camera coordinates. In contrast, we propose a more robust approach based on 3D Gaussian Splatting. Since the predicted Gaussians are pixel-aligned with the input image and located in the coordinate frame of the first image, they can be treated as being in the camera coordinate system. Thus, the unknown parameters are the intrinsic matrix, which can be optimized via the formulation:

\begin{equation}
    \mathbf{K}^* = \mathop{\min}_{\mathbf{K}} \sum_{u=0}^W \sum_{v=0}^{H}  \mathcal{L}_{c_1}
    \label{eq:intrisic_opt}
\end{equation}

where $\mathcal{L}_{c_1} = \left| I_{1}^{u,v}(\mathbf{K}, \mathcal{G}) - \bar{I}_1^{u,v} \right|_1$, $\bar{I}_1^{u,v}$ represents the observed pixel value at position $(u,v)$ in the input image, and $I(\mathbf{K}, \mathcal{G})$ renders the image based on the intrinsic parameters $\mathbf{K}$ and the initial Gaussian surfels $\mathcal{G}$. \( W \) and \( H \) represent the image width and height. Since the Gaussian primitives are defined in canonical coordinates, the intrinsic parameters can be optimized by minimizing Equation~\ref{eq:intrisic_opt} while fixing the Gaussians. 

The differential process for optimizing the intrinsic parameters is implemented within the Gaussian Splatting module, leveraging the CUDA library for efficient computation. The terms in Equation~\ref{eq:intrisic_opt} are differentiable with respect to the camera intrinsic matrix $\mathbf{K}$ via the chain rule:
\begin{equation}
    \begin{split}
        \frac{\partial \bm{\mu}_I}{\partial \mathbf{K}} &= \frac{\partial \bm{\mu}_I}{\partial \bm{\mu}_C} \frac{\partial \bm{\mu}_C}{\partial\mathcal{P}}\frac{\partial \mathcal{P}}{\partial \mathbf{K}} \\
        \frac{\partial \bm{\Sigma}_I}{\partial \mathbf{K}} &= \frac{\partial \bm{\Sigma}_I}{\partial \mathbf{J}} \frac{\partial \mathbf{J}}{\partial \mathbf{K}}
    \end{split}
\end{equation}
where $\mathbf{J}$ is the Jacobian of the affine approximation of the projective transformation $\mathcal{P}$~\cite{kerbl20233d}, $\bm{\mu}_I$ and $\bm{\Sigma}_I$ are the 2D Gaussian on the image plane, while $\bm{\mu}_C$ and $\bm{\Sigma}_C$ are 3D Gaussian in the canonical coordinates. 
The detailed derivation of the gradients with respect to the camera intrinsic matrix $\mathbf{K}$ is provided in the supplementary materials (see Section~\ref{subsec:Projection_Process} and \ref{subsec:affine_approx}).

To assist with the convergence of the optimization process, we initialize the principal point at the center of the image plane, with \( c_x = 0.5W \) and \( c_y = 0.5H \), respectively. The initial focal length is set to \( 1.2 \) times the image dimensions (width and height) to provide a reasonable starting estimate for the camera's optical properties.
It is important to note that the offset of the principal point is relatively small compared to the amount of adjustment needed for the camera's focal length. Therefore, we use separate learning rates, \( l_c \) for the principal point parameters and \( l_f \) for the focal length, to ensure efficient and stable optimization.

\paragraph{Extrinsic Parameter Estimation.}

The camera extrinsic matrix \( \mathbf{T} = \left[ \begin{array}{cc}
    \mathbf{W} & \mathbf{t}  \\
    0 & 1 
\end{array}\right] \in SE(3) \) consists of the rotation matrix \( \mathbf{W} \in SO(3) \) and the translation vector \( \mathbf{t} \in \mathbb{R}^3 \), which describes the rigid transformation between the two camera views, \( I_1 \) and \( I_2 \). In traditional camera pose estimation methods~\cite{mur2015orb, schoenberger2016mvs}, the common approach is to re-project 3D point clouds from world coordinates back into the image plane for camera tracking. However, in this section, we introduce the strategy for camera pose estimation based on the predicted Gaussian surfels. 

To optimize the relative camera pose between the first and second views, we fix the intrinsic parameters and the Gaussians predicted by the first image, and iteratively refine the second view's pose in canonical coordinates. The optimization is performed by minimizing the loss function defined as:
\[
\mathbf{T}^{*} = \mathop{\min}\limits_{\mathbf{T}} \sum_{u=0}^{W} \sum_{v=0}^{H} \mathcal{L}_{c_2}
\label{eq:extrisic_opt}
\]
where \( \mathcal{L}_{c_2} = | I^{u,v}_{2}(\mathbf{K}, \mathcal{G}, \mathbf{T}) - \bar{I}^{u,v}_{2} |_1 \), and \( \bar{I}_{2} \) is the observed image. The optimization of the camera extrinsics proceeds by calculating the gradients of the error term with respect to the camera pose parameters. The derivatives of the 2D Gaussian position \( \bm{\mu}_I \) and covariance matrix \( \bm{\Sigma}_I \) with respect to the camera extrinsics \( \mathbf{T} \) are computed as follows:

\begin{equation}
\begin{split}
    \frac{\partial \bm{\mu}_{I}}{\partial \mathbf{T}} &= \frac{\partial \bm{\mu}_{I}}{\partial \bm{\mu}_C} \frac{\partial \bm{\mu}_C}{\partial \mathbf{T}} \\
    \frac{\partial \bm{\Sigma}_{I}}{\partial \mathbf{T}} &= \frac{\partial \bm{\Sigma}_{I}}{\partial \mathbf{J}} \frac{\partial \mathbf{J}}{\partial \bm{\mu}_C} \frac{\partial \bm{\mu}_C}{\partial \mathbf{T}} + \frac{\partial \bm{\Sigma}_{I}}{\partial \mathbf{W}} \frac{\partial \mathbf{W}}{\partial \mathbf{T}}
\end{split}
\end{equation}
where \( \mathbf{T} \) is the camera extrinsic matrix, and the gradient update for the parameters \( \mathbf{T} \) occurs on the manifold \( \mathfrak{se}(3) \) via \textit{Lie} algebra representations~\cite{matsuki2023gaussian}.

\subsection{Bundle-Adjusting Gaussian Splatting}

After obtaining the initial camera intrinsic parameters and relative camera poses (see Section~\ref{sec:camera_para_op}), the Gaussian surfels are rasterized into corresponding depth maps \( D_{1} \) and \( D_{2} \) using the alpha-blending algorithm. The depth map \( D_{2} \) from the second viewpoint is transformed to the first viewpoint using the following warping operation:
\begin{equation}
    D_{1,warp} = \Pi( D_{2}, \mathbf{K}, \mathbf{T} )
\end{equation}
where \( \Pi(\cdot) \) represents the warp operation that projects the depth map \( D_{2} \) from the second camera view to the first camera view using the estimated intrinsic matrix \( \mathbf{K} \) and the relative camera pose \( \mathbf{T} \).

Then, a geometric constraint is established to enforce better multi-view consistency of the Gaussian radiance fields in terms of geometry. This constraint is formulated as follows:
\begin{equation}
    \mathcal{L}_{geo} = | D_{1,warp} - D_{1} |.  
\end{equation}

And the photometric error between rendered and observed images is defined as 
\begin{equation}
   \mathcal{L}_{pho} = (1.0 - \lambda_{sm})\mathcal{L}_{c_i} + \lambda_{sm} |1.0 - \mathcal{L}_{{ssim}_{i}} |
\end{equation}
where $i \in [1,2]$, and $\mathcal{L}_{{ssim}_{i}} =ssim(I^{u,v}_{i} - \bar{I}^{u,v}_i)$ is to the compute structural similarity index measure distance between two images.   

To jointly optimize the scaled radiance fields, including Gaussian surfels, as well as the camera intrinsic and extrinsic parameters, for the input sparse views, we define a comprehensive loss function to supervise the refinement process:

\begin{equation}
\mathcal{G}^{*}, \mathbf{K}^{*}, \mathbf{T}^{*} = \mathop{\min}\limits_{\mathcal{G}, \mathbf{K}, \mathbf{T}} \sum_{u=0}^W \sum_{v=0}^H \left( \lambda_1 \mathcal{L}_{pho_1} + \lambda_2 \mathcal{L}_{pho_2} + \lambda_3 \mathcal{L}_{geo} \right)
\end{equation}
where \( \lambda_1 \), \( \lambda_2 \), and \( \lambda_3 \) are weighting parameters that balance the contributions of the respective loss terms. 
This joint optimization allows for the simultaneous refinement of the Gaussian surfels, camera intrinsics, and extrinsics, improving the overall performance of the sparse-view radiance fields reconstruction. 

To clarify the refinement stage, we illustrate the strategy of this stage used in the experiments, as shown in Algorithm~\ref{alg:opti}. The following procedure outlines the iterative process for intrinsics, Gaussian primitives, extrinsics, and joint optimization:

\begin{algorithm}
\caption{Strategy for achieving scaled Gaussian radiance fields}\label{alg:opti}
    \begin{algorithmic}[1]
    \REQUIRE $\mathcal{G}$, $I^{c_1}$, and $I^{c_2}$ 
        \ENSURE Iteration $n = 100$
        \FOR{$j = 0$ to $10$}
        \STATE     $\mathbf{K}$ \; optimization
        \ENDFOR
        \FOR{$j = 10$ to $20$}
        \STATE     $\mathcal{G} \; optimization$ 
        \ENDFOR
        \FOR{$j = 20$ to $40$}
        \STATE     $\mathbf{T}  \; optimization$ 
        \IF {$\mathbf{T} \to \mathbf{T}^*$}
        \STATE $Finish $
        \ENDIF 
        \ENDFOR
        \FOR{$j = 40$ to $100$}
            \STATE $\mathcal{G}, \mathbf{K}, \mathbf{T} \;optimization$ 
        \ENDFOR
    \end{algorithmic}
\end{algorithm}

\section{Experiments}
\label{sec:experi}
In this section, we present both quantitative and qualitative evaluations comparing the proposed method with state-of-the-art algorithms, focusing on performance in the novel view rendering task.


\subsection{Implementation Details and Baselines}
SmileSplat consists of two main modules: Gaussian prediction and radiance field bundle adjustment. For the first module, which handles Gaussian surfel prediction, we utilize a neural network built on the PyTorch library. The encoder is based on a Vision Transformer (ViT) model with a patch size of 16, while the decoder is based on a ViT-base architecture. This module includes four separate heads for Gaussian parameter registration, designed to generate Gaussians for an image resolution of $224 \times 224$. The specific size and scale of each head are provided in the supplementary material.
The second module, radiance field bundle adjustment, incorporates both intrinsic and extrinsic parameters. This module is implemented using the CUDA library to leverage GPU acceleration for efficient computations.
For training, we use two GeForce RTX 3090 GPUs. During the inference stage, only a single GPU is used in our experiments.
Throughout all experiments, we maintain a consistent learning rate of 0.0002 for Gaussian optimization. The weighting $\lambda_{ssim}$ for the photometric loss is 0.6. Additionally, the weights $\lambda_1$, $\lambda_2$, and $\lambda_3$ are set to 0.05, 0.05 and 0.01, respectively.


The state-of-the-art approaches are compared against with the proposed method in the novel view synthesis task, which are categeried as two types, where the first set of methods, including pixelNeRF~\cite{yu2021pixelnerf}, AttnRend~\cite{du2023learning}, pixelSplat~\cite{charatan2024pixelsplat}, and MVSplat~\cite{chen2024mvsplat}, needs camera parameters in their training and testing process, while the second set of approaches, DUSt3R~\cite{wang2023dust3r}, MASt3R~\cite{mast3r_arxiv24}, Splatt3R~\cite{smart2024splatt3r}, and NoPoSplat~\cite{ye2024no}, have capbilities to estimate intrisic and extrinsic. 

\subsection{Datasets and Metrics}
Following the experimental setup of pioneering approaches~\cite{charatan2023pixelsplat, chen2024mvsplat} in this domain, we evaluate the proposed method on two large-scale datasets. The first dataset, RealEstate10K (Re10K)~\cite{zhou2018stereo}, is collected from real-estate sequences on YouTube. Based on the training/testing split, the dataset contains 29,144 scenes for training and 7286 scenes for testing, respectively. The second dataset, ACID\cite{liu2021infinite}, focuses on nature scenes collected from the viewpoints of aerial drones. To evaluate the reconstruction capabilities of the proposed method, we further assess the trained model across additional datasets, including Replica~\cite{straub2019replica} and ICL-NUIM~\cite{handa2014benchmark}.

For quantitative evaluation, we report the rendering performance using standard image quality metrics, including Peak Signal-to-Noise Ratio (PSNR), Structural Similarity Index Measure (SSIM), and Learned Perceptual Image Patch Similarity (LPIPS). Specifically, the first two metrics (PSNR and SSIM) evaluate the color-wise similarity and structural similarity between the rendered and observed images. The third metric, LPIPS, compares the feature-level similarity between two images, using features extracted by a pre-trained neural network (e.g., VGG-Net~\cite{simonyan2014very}) rather than directly comparing the pixel values of the images.
To analyze the relationship between rendering performance and the overlap of input images, the visual overlap $\gamma$ between two input images is computed based on a dense feature matching method~\cite{edstedt2024roma}. The overlap is then classified into three levels: Small ($\gamma\leq 0.3\%$), Medium ($0.3\%\leq \gamma \leq 0.55\% $, and Large ($\gamma \geq 0.55\%$), following the method of~\cite{ye2024no}.



\subsection{Novel View Rendering}

\begin{figure*}
    \centering
    \includegraphics[width=\textwidth]{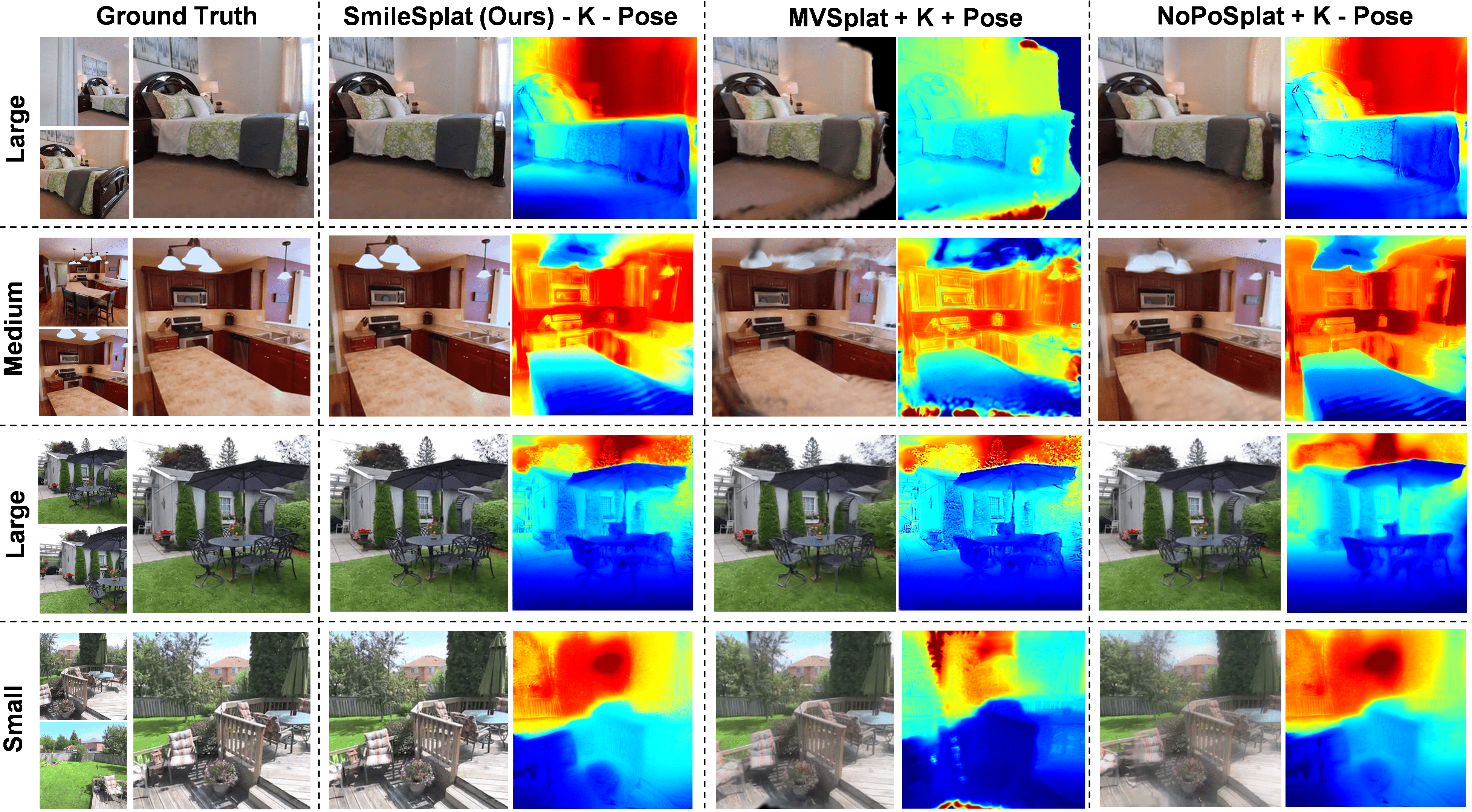}
    \caption{Comparison of novel view rendering and depth prediction results on Re10k~\cite{zhou2018stereo} and ACID~\cite{liu2021infinite} datasets with different settings. The rendering results include the RGB in the left column and the depth in the right column. $\pm$ K and $\pm$ Pose denote whether intrinsics and extrinsics are free or not.}
    \label{fig:render_re10k}
\end{figure*}

\begin{table*}[]
\resizebox{\textwidth}{!}{
\begin{tabular}{cc|cccc|ccccc}
\toprule
\multicolumn{2}{c|}{Method}       & pixelNeRF~\cite{yu2021pixelnerf} & AttnRend~\cite{du2023learning} & pixelSplat~\cite{charatan2024pixelsplat} & MVSplat~\cite{chen2024mvsplat} & DUSt3R~\cite{wang2023dust3r} & MASt3R~\cite{mast3r_arxiv24} & Splatt3R~\cite{smart2024splatt3r} & NoPoSplat~\cite{ye2024no} & SmileSplat \\ \hline
\multirow{2}{*}{Setting} & K     &  $\checkmark$   & $\checkmark$  & $\checkmark$        & $\checkmark$         & $\times$   &$\times$    &$\times$    & $\checkmark$   & $\times$   \\
    & Pose  &  $\checkmark$   & $\checkmark$  & $\checkmark$        & $\checkmark$         & $\times$   &$\times$    &$\times$    & $\times$   & $\times$            \\  \hline
\multirow{3}{*}{Small}   & PSNR $\uparrow$ 
&18.41   & 19.15  & 20.26 & 20.35  &14.10 & 13.53  &14.35  & 22.51 & \textbf{26.81}      \\
                         & SSIM $\uparrow$ 
&0.601   &0.663   &0.717  &0.724   &0.432 &0.407  &0.475   & 0.585 &  \textbf{0.856}    \\ 
                         & LPIPS $\downarrow$ 
& 0.526  &0.368   &0.266  &0.250   &0.468 &0.494  &0.472   &0.462  & \textbf{0.071}    \\ \hline
\multirow{3}{*}{Medium}  & PSNR$\uparrow$  
&19.93  & 22.53 & 23.71 & 23.77  & 15.41 & 14.94  &15.52 &24.89 & \textbf{26.60 }      \\
                         & SSIM$\uparrow$  
&0.632  & 0.763 &0.809  &0.812   &0.451  &0.436   &0.502 &0.616 &  \textbf{0.843}      \\ 
                         & LPIPS $\downarrow$ 
&0.480  & 0.269 & 0.181 & 0.173  & 0.432 & 0.451  &0.425 & 0.160 & \textbf{0.071}    \\ \hline
\multirow{3}{*}{Large}   & PSNR$\uparrow$  
& 20.86  &25.89   & 27.15  & 27.40   &16.42   & 16.02   & 15.81  & \textbf{27.41}  & 27.26           \\
                         & SSIM$\uparrow$  
& 0.639  & 0.845  & 0.879  &0.884    &0.453   &0.444    &0.483   &\textbf{0.883}   & 0.850           \\
                         & LPIPS $\downarrow$ 
& 0.458  & 0.186  & 0.122  &0.116   & 0.402   &  0.418  &0.421   &0.119   & \textbf{0.063}        \\ \hline
                         \bottomrule
\end{tabular}}
\caption{Quantitative comparisons of novel view rendering on the Re10K~\cite{zhou2018stereo}. $\checkmark$ and $\times$ in the setting rows show whether corresponding camera parameters are required or not during the inference.}
\label{tab:rendering_re10k}
\end{table*}

As shown in Table~\ref{tab:rendering_re10k}, state-of-the-art methods, both CamPara-Free and CamPara-Required, are compared with the proposed approach, SmileSplat, in the novel view rendering task. For CamPara-required methods, Gaussian Splatting approaches, such as pixelSplat and MVSplat, demonstrate superior rendering quality compared to implicit representation methods like pixelNeRF.
For CamPara-free methods, DUSt3R~\cite{wang2023dust3r} and MASt3R~\cite{mast3r_arxiv24} predict pixel-aligned point clouds based on the input images, and other intrinsic and extrinsic parameters can be estimated and optimized using conventional multi-view geometry algorithms. Building upon the architectures of these methods, the approaches Splatt3R~\cite{smart2024splatt3r} and NoPoSplat~\cite{ye2024no} introduce additional heads to estimate the parameters of 3D Gaussian ellipsoids, which significantly improve rendering performance from $14.49$ to $23.08$ for inputs with small visual overlaps. Unlike the Re10K dataset, which consists of real-estate sequences, the ACID dataset focuses on natural scenarios. However, the trends observed in Table~\ref{tab:rendering_re10k} and Figure~\ref{fig:render_re10k} can also be seen in the supplementary.

\begin{figure*}
    \centering
    \includegraphics[width=\textwidth]{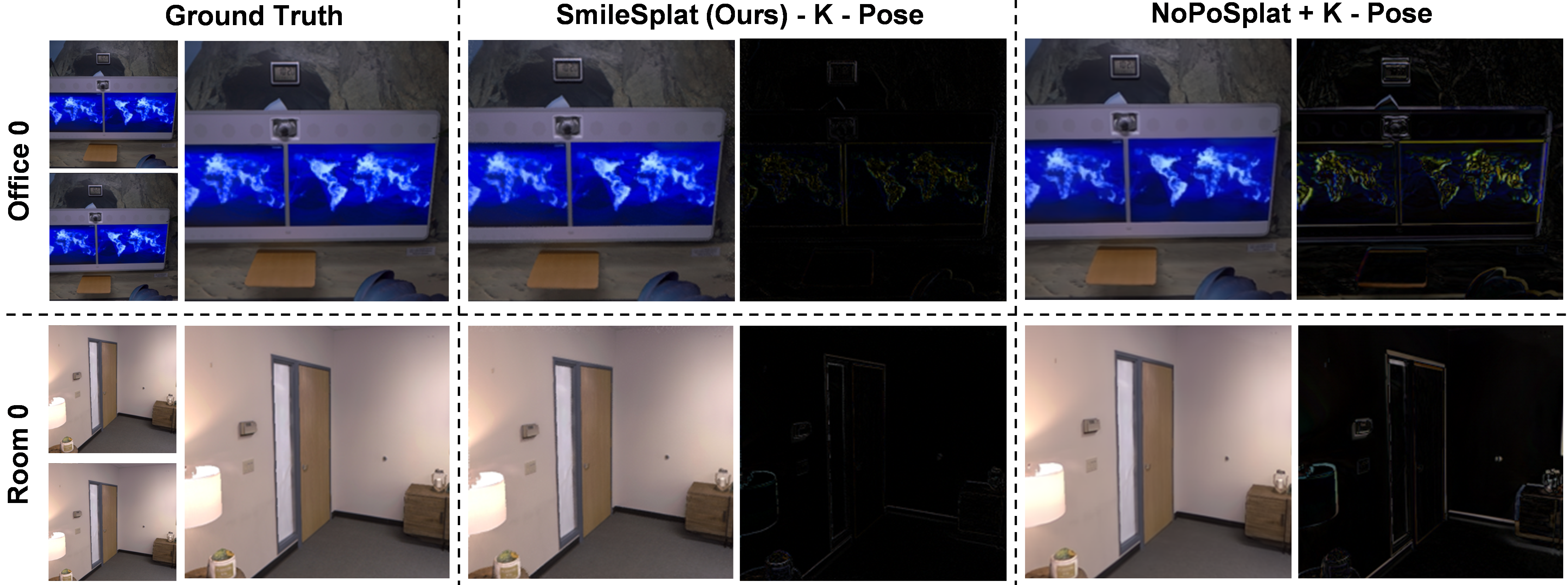}
    \caption{Comparisons of novel view rendering on Replica~\cite{straub2019replica} dataset. The rendering results include the RGB in the left column and the difference with the ground truth in the right column. $\pm$ K and $\pm$ Pose denote whether intrinsics and extrinsics are free or not.}
    \label{fig:cross-dataset-replica}
\end{figure*}

\begin{table*}
\centering
\resizebox{\linewidth}{!}{
\small
\begin{tabular}{l|ccc|ccc|ccc}
    \toprule
    \multirow{2}{*}{Method}  & \multicolumn{3}{c|}{View 1} &\multicolumn{3}{c|}{View 2} &\multicolumn{3}{c}{View 3} \\
         & PSNR $\uparrow$    &SSIM$\uparrow$  &LPIPS$\downarrow$    & PSNR $\uparrow$    &SSIM$\uparrow$  &LPIPS$\downarrow$  & PSNR $\uparrow$    &SSIM$\uparrow$  &LPIPS$\downarrow$   \\
    \midrule
    $\mathcal{P}_{h}$+$Cam_{a}$   &23.98 &0.769 &0.227 &15.80 & 0.486 &0.355 &18.63 &0.582 &0.296  \\
    $\mathcal{P}_{h}$+$Cam_{b}$  & 23.67 &0.763 &0.183  &16.72 &0.538 &0.283 &17.85 &0.559 &0.266 \\ \hline
    $\mathcal{G}_h$ + $Cam_{a}$ + $\mathcal{L}_{pho}$: $\mathcal{O}_{\mathcal{G}}$  &28.43 &0.911 &0.083 &18.61 &0.743 &0.234 &18.63 &0.744 &0.234 \\
    $\mathcal{G}_h$ + $Cam_{b}$ + $\mathcal{L}_{pho}$: $\mathcal{O}_{\mathcal{G}}$ &29.06&0.927&0.070 &20.64 &0.807 &0.188 & 19.27 &0.608 &0.240  \\ \hline
    \bottomrule
\end{tabular}}
\resizebox{0.49\linewidth}{!}{
\begin{tabular}{c|lll}
\toprule
\multirow{5}{*}{$\rotatebox{90}{Initial}$}  &$\mathcal{P}_{h}$  & Pointmap              & point cloud prediction  of Dust3R   \\
& $\mathcal{G}_h$ &$[\mu \; \mathbf{n} \; o \; \mathbf{s}]$  & Gaussian representation prediction of our method \\  \cline{2-4} 
& $Cam_{a}$    & K, R, t           & Intrinsics and extransics  of Dust3R           \\
 & $Cam_{b}$   & K, R, t           & Intrinsics and extransics  of ICP   
\\   \bottomrule     
\end{tabular}}
\resizebox{0.5\linewidth}{!}{
\begin{tabular}{c|lll}
\toprule
\multicolumn{1}{l|}{\multirow{3}{*}{$\rotatebox{90}{\small Optimization}$ }} & $\mathcal{L}_{pho}$   & L\_\{pho\}        & photometric constraint between RGB images     \\  \cline{2-4} 
\multicolumn{1}{l|}{}    & $\mathcal{O}_{\mathcal{G}}$  & $[\bm{\mu} \; \mathbf{n} \; \mathbf{s} \; o]$            & Gaussian surfel optimization       \\    
\multicolumn{1}{l|}{}    & $\mathcal{O}_{K}$   &   $f_x, f_y, c_z, c_y$ & intrinsic optimization       \\             
\multicolumn{1}{l|}{}    & $\mathcal{O}_{pose}$  & \textit{Lie} algebra  & camera pose optimization       \\     
\bottomrule     
\end{tabular}}
\caption{Ablation studies for testing different modules in the novel view rendering task based on the Re10K~\cite{zhou2018stereo} dataset. For Gaussian prediction testing, all three views are novel views, while View 3 is the novel view when the first two views are used for reference Gaussian surfels. More settings are provided in Supplementary.}
\label{tab:ablation_study}
\end{table*}




\subsection{Cross-Dataset Generalization.} 
As shown in Table~\ref{tab:cross-dataset-replica}, both NoPoSplat and the proposed SmileSplat demonstrate superior robustness and accuracy compared to other CamPara-Free and CamPara-Required approaches. In this section, we continue to evaluate the zero-shot performance of NoPoSplat and SmileSplat by directly applying their models to the Replica sequences~\cite{straub2019replica}. It is important to note that both models were trained on the Re10K dataset without any further fine-tuning on the Replica dataset.

Table~\ref{tab:cross-dataset-replica} lists four sequences, including \textit{Office 0}, \textit{Office 1}, \textit{Room 0}, and \textit{Room 1}, which are used to compare the novel view rendering performance. Similar to the settings used in the Re10K testing, two images are fed into the systems for Gaussian radiance field reconstruction, and a novel view is used for testing. Based on the number of input images, the sequences are categorized into three classes: Small (7 images), Medium (12 images), and Large (20 images). Within each class, the proposed SmileSplat consistently robustness performance.
Specifically, in the \textit{Room 1} sequence, the PSNR result for NoPoSplat is $27.64$, which is improved by $25\%$ to $34.60$ with our method. However, when directly comparing the rendering quality of NoPoSplat, as shown in Figure~\ref{fig:cross-dataset-replica}, the rendered images still exhibit high fidelity. To better understand the gap between the quantitative and qualitative results, we compute the photometric distance between the rendered images and the corresponding ground truth images, as shown in Figure~\ref{fig:cross-dataset-replica}. The results reveal that the alignment between the rendered images and the ground truth images has notable issues, especially in the \textit{Office} sequence. In textured regions, the photometric error between the rendered and ground truth images is significantly higher for NoPoSplat. Since the camera pose of the reference image is estimated by these methods themselves, the observed phenomenon suggests that the proposed SmileSplat method achieves better alignment and overall performance than NoPoSplat. More detailed camera pose experiments are provided in the supplementary materials.

\begin{table}
    \resizebox{\linewidth}{!}{
    \begin{tabular}{cc|cccc|cccc}
    \toprule
       \multirow{2}{*}{Method}  & \ & \multicolumn{4}{c|}{NoPoSlpat~\cite{ye2024no}} & \multicolumn{4}{c}{SmileSplat (Ours)} \\ 
       & \ & Small & Medium & Large & Ave. & Small & Medium & Large & Ave. \\ 
    \midrule
         \multirow{3}{*}{Office 0} & PSNR & 32.64  & 28.61 & 30.99 & 30.44  & \textbf{35.62} & \textbf{33.82} & \textbf{31.65} & \textbf{33.71}\\
         & SSIM & 0.878  & 0.845 & 0.869 & 0.861  & \textbf{0.953} & \textbf{0.942} & \textbf{0.923} & \textbf{0.940} \\
         & LPIPS & 0.166  & 0.164 & 0.172 & 0.167  & \textbf{0.036} & \textbf{0.049} & \textbf{0.071} & \textbf{0.052} \\
    \midrule
         \multirow{3}{*}{Office 1} & PSNR & 32.65  & 28.59 & 30.95 & 30.42  & \textbf{34.22} & \textbf{33.27} & \textbf{31.79} & \textbf{33.39}\\
         & SSIM & 0.877  & 0.844 & 0.869 & 0.861  & \textbf{0.934} & \textbf{0.926} & \textbf{0.918} & \textbf{0.928}\\
         & LPIPS & 0.166  & 0.164 & 0.172 & 0.167  & \textbf{0.056} & \textbf{0.062} & \textbf{0.077} & \textbf{0.063}\\
    \midrule
         \multirow{3}{*}{Room 0} & PSNR & 28.43  & 30.37 & 32.60 & 30.47  & \textbf{31.11} & \textbf{31.76} & \textbf{30.89} & \textbf{31.33} \\
         & SSIM & 0.684  & 0.723 & 0.723 & 0.702  & \textbf{0.894} & \textbf{0.913} & \textbf{0.895} & \textbf{0.902} \\
         & LPIPS & 0.157  & 0.150 & 0.150 & 0.155  & \textbf{0.025} & \textbf{0.016} & \textbf{0.024} & \textbf{0.021}\\
    \midrule
         \multirow{3}{*}{Room 1} & PSNR & 27.64  & 27.73 & 29.50 & 28.29  & \textbf{34.60} & \textbf{33.64} & \textbf{32.39} & \textbf{33.83}\\
         & SSIM & 0.836  & 0.134 & 0.729 & 0.780  & \textbf{0.961} & \textbf{0.956} & \textbf{0.949} & \textbf{0.957}\\
         & LPIPS & 0.133  & 0.777 & 0.140 & 0.135  & \textbf{0.015} & \textbf{0.019} & \textbf{0.025} & \textbf{0.018}\\
    \bottomrule
    \end{tabular}}
    \caption{Cross-Dataset generalization tests on the Replica~\cite{straub2019replica} dataset.}
    \label{tab:cross-dataset-replica}
\end{table}

\subsection{Ablation Study}


In this section, we analyze the performance of different modules proposed in our method. As shown in Table~\ref{tab:ablation_study}, various settings of modules are integrated and tested on the Re10K dataset benchmark.
First, we render the predicted Gaussians at three different viewpoints using two different initial camera parameter estimation methods. Since the Gaussian parameters are predicted at the coordinates of View 1, camera pose errors result in degraded performance when rendering from the other two viewpoints.
When we optimize the Gaussian parameters based on feedback from the photometric loss of View 1 and View 2, the rendering quality for these two views improves significantly. However, this optimization does not have a strong positive impact on the novel view rendering (View 3), as the camera pose of View 3 cannot be accurately estimated. This suggests that while feedback from a few views helps improve performance for those specific viewpoints, the lack of accurate pose estimation for novel views limits the overall benefit.






\section{Discussion and Conclusion}
In this paper, we introduce a robust and generalizable Gaussian Splatting method for unconstrained sparse images. To improve multi-view consistency in 3D Gaussian models, we estimate the Gaussian surfel parameters using our multi-head Gaussian registration framework. To avoid reliance on ground truth camera intrinsics and extrinsics, these parameters are optimized directly in the canonical coordinate frame based on the bundle-adjusting Gaussian Splatting module, to achieve accurate scaled Gaussian radiance fields. 
\appendix

\section{Fundamental Theory in RGB and Depth Rendering}

\subsection{Novel View Rendering}
During the Gaussian Splatting optimization process, the rendered RGB and depth maps used in Section~\textcolor{red}{3.2}, are computed based on the color $\mathbf{c}_i$ (or depth) and opacity $\alpha_i$ of the involved Gaussians $\mathcal{G}_i$:

\begin{equation}
    \left\{
    \begin{split}
         \mathcal{C}_p & = \sum_{i\in N} c_i \alpha_i \prod_{j=1}^{i-1} (1-\alpha_j) \\
        \mathcal{D}_p  & = \sum_{i\in N} z_i \alpha_i \prod_{j=1}^{i-1} (1-\alpha_j) \\
    \end{split} \right.
\end{equation}
where $z_i$ is the distance to the position of the Gaussian surfel along the camera ray, and the $\alpha_i$ can be computed based on the 2D covariance $\bm{\Sigma}_I \in \mathbb{R}^{2 \times 2}$ and the opacity value $o_i$ following the 3DGS algorithm~\cite{kerbl20233d}.




\subsection{Projection Process in Gaussian Splatting}
\label{subsec:Projection_Process}

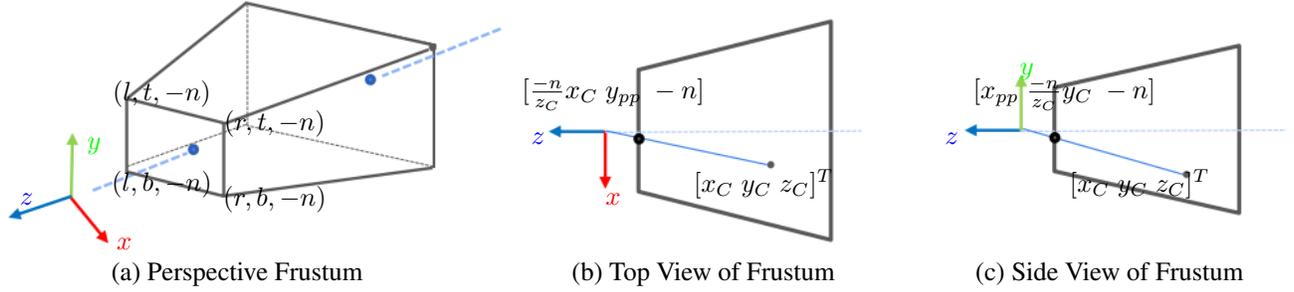
\begin{figure*}
\begin{tikzpicture}[spy using outlines={red,magnification=3,size=2cm}, connect spies] 
\node (img1) at (0,0){ \includegraphics[width=\linewidth]{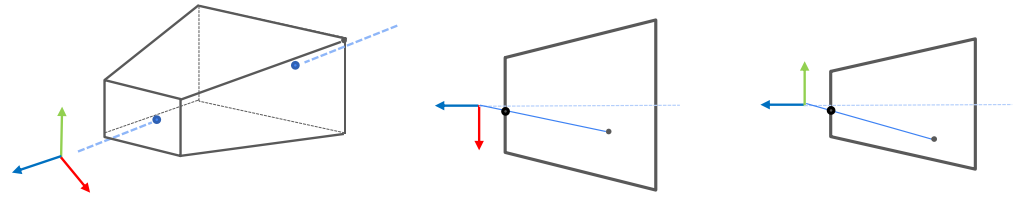}};
\node[] at (-7, -1.6) {\textcolor{red}{$x$}};
\node[] at (-7.4, -0.3) {\textcolor{green}{$y$}};
\node[] at (-8.3, -1.0) {\textcolor{blue}{$z$}};
\node[] at (-5.5,-2) {(a) Perspective Frustum};
\node[] at (-6.5, -0.8) {$(l, b, -n)$}; \node[] at (-6.5, 0.4) {$(l, t, -n)$};
\node[] at (-5.0, -1.0) {$(r, b, -n)$}; \node[] at (-5.0, 0) {$(r, t, -n)$};
\node[] at (0.7,-2) {(b) Top View of Frustum};
\node[] at (1.5, -0.8) {$[x_C\; y_C\; z_C]^T$}; \node[] at (-0.5, 0.4) {$[\frac{-n}{z_C}x_C \; y_{pp} \; -n ]$};
\node[] at (-0.5, -1.0) {\textcolor{red}{$x$}};
\node[] at (-1.5, -0.2) {\textcolor{blue}{$z$}};
\node[] at (6.1,-2) {(c) Side View of Frustum};
\node[] at (6.5, -0.8) {$[x_C\; y_C\; z_C]^T$}; \node[] at (5.5, 0.4) {$[x_{pp} \; \frac{-n}{z_C}y_C \; -n ]$};
\node[] at (5.0, 0.7) {\textcolor{green}{$y$}};
\node[] at (4.0, -0.2) {\textcolor{blue}{$z$}};
\end{tikzpicture}
    \caption{3D Perspective frustum and its different views in 2D space.}
    \label{fig:projective_matrix}
\end{figure*}

In perspective projection, the 3D position $\bm{\mu}_C$ (the mean vector of the Gaussian in camera coordinates) within a truncated pyramid frustum, as illustrated in Figure~\ref{fig:projective_matrix}(a), can be mapped to normalized device coordinates (NDC) through the following two steps: (1) projecting $\bm{\mu}_C$ onto the projection plane, and (2) mapping the projection onto the normalized device coordinates. 

Based on the top view, as shown in Figure~\ref{fig:projective_matrix}(b), of the truncated pyramid frustum, the following relationship can be derived:
\begin{equation}
\left\{
\begin{split}
    x_{pp} &= -\frac{n}{z_C} x_C \\
    y_{pp} &= -\frac{n}{z_C} y_C
\end{split}
\right.
\label{eq:projective_step_1}
\end{equation}
where \(x_{pp}\) and \(y_{pp}\) represent the coordinates on the projection plane, and \(\bm{\mu}_C = [x_C \; y_C \; z_C]^T\) denotes the 3D position in camera coordinates. In the truncated pyramid frustum, the range of the \(x\)-coordinate spans from \(l\) to \(r\), while the ranges for the \(y\)- and \(z\)-coordinates are \([b, t]\) and \([-n, -f]\), respectively.


When mapping the range \([l \to r]\) of the projection plane in the \(x\)-direction to the normalized device coordinates (NDC) range \([-1 \to 1]\), the mapping function can be defined as a linear transformation:

\begin{equation}
    x_{ndc} = a x_{pp} + b    
\end{equation}
where \(a = \frac{1 - (-1)}{r - l}\) and \(b\) can be computed as \(b = 1 - \frac{2r}{r - l}\) when \(x_{pp}\) is set to \(r\). Therefore, the mapping relationship between \(x_{pp}\) and \(x_{ndc}\) is given by:
\begin{equation}
    x_{ndc} = \frac{2}{r - l} x_{pp} - \frac{r + l}{r - l}.
\label{eq:x_ndc}
\end{equation}

Similarly, the mapping relationship in the \(y\)-direction can be expressed as:

\begin{equation}
    y_{ndc} = \frac{2}{t - b} y_{pp} - \frac{t + b}{t - b}
  \label{eq:y_ndc}
\end{equation}
where \(t\) and \(b\) represent the upper and lower bounds of the projection plane in the \(y\)-direction, respectively, as shown in Figure~\ref{fig:projective_matrix}.



By substituting \([x_C \; y_C]\) from Equation~\ref{eq:projective_step_1} in place of \([x_{pp} \; y_{pp}]\) in Equations~\ref{eq:x_ndc} and~\ref{eq:y_ndc}, the mapping function from camera coordinates to NDC can be expressed as:

\begin{equation}
    \left[\begin{array}{c}
         x_{ndc}  \\
         y_{ndc}  \\
    \end{array}\right] = \left[\begin{array}{c}
        -\frac{2}{r-l} \frac{n}{z_C} x_C - \frac{r+l}{r-l} \\
        -\frac{2}{t-b} \frac{n}{z_C} y_C - \frac{t+b}{t-b} \\
    \end{array}\right]
\end{equation}
which can then be simplified as functions of
\begin{equation}
    \left[\begin{array}{c}
         x_{ndc}  \\
         y_{ndc}  \\
    \end{array}\right] = \left[\begin{array}{c}
        -(\frac{2n}{r-l} x_C + \frac{r+l}{r-l}z_C) \frac{1}{z_C} \\
        -(\frac{2n}{t-b} y_C + \frac{t+b}{t-b} z_C) \frac{1}{z_C} \\
    \end{array}\right].
    \label{eq:ndc_xyc}
\end{equation}



By setting $w_{clip}=- z_C$, the Equation~\ref{eq:ndc_xyc} can be rewritten as:
\begin{equation}
    \begin{split}
    \left[ \begin{array}{c}
         x_{ndc}   \\
         y_{ndc} \\
         z_{ndc}
    \end{array} \right] &=  \left[ \begin{array}{c}
         x_{clip}/w_{clip}   \\
         y_{clip}/w_{clip} \\
         z_{clip}/w_{clip}
    \end{array} \right] 
    \end{split}
\end{equation}

\begin{equation}
    \begin{split}
       \left[ \begin{array}{c}
         x_{clip}  \\
         y_{clip}  \\
         z_{clip} \\
         w_{clip} \\
    \end{array} \right]  & =  \left[\begin{array}{c}
        \frac{2n}{r-l} x_C + \frac{r+l}{r-l}z_C  \\
        \frac{2n}{t-b} y_C + \frac{t+b}{t-b} z_C  \\
       \frac{-(f+n)}{f-n} z_C +\frac{-2fn}{f-n}  \\
       -z_C
    \end{array}\right] \\
       &=  \left[\begin{array}{cccc}
        \frac{2n}{r-l} & 0 & \frac{r+l}{r-l} &0  \\
        0 & \frac{2n}{t-b} & \frac{t+b}{t-b} &0 \\
       0  &0 &\frac{-(f+n)}{f-n} &\frac{-2fn}{f-n} \\
       0 & 0 & -1 & 0
    \end{array}\right] \left[ \begin{array}{c}
         x_{C}   \\
         y_{C} \\
         z_{C} \\
         1
    \end{array} \right]
    \end{split}
    \label{eq:clip_C}
\end{equation}
 which can be denoted in a more familiar form as:
 
\begin{equation}
    \bm{\mu}_{clip} = \mathcal{P}[\bm{\mu}_C^T \; 1]^T
\end{equation}
where $\bm{\mu}_{clip} = [x_{clip} \; y_{clip} \;z_{clip} \;w_{clip}]^{T}$ and $\mathcal{P}$ is represented by $\left[\begin{array}{cccc}
        \frac{2n}{r-l} & 0 & \frac{r+l}{r-l} &0  \\
        0 & \frac{2n}{t-b} & \frac{t+b}{t-b} &0 \\
       0  &0 &\frac{-(f+n)}{f-n} &\frac{-2fn}{f-n} \\
       0 & 0 & -1 & 0
    \end{array}\right]$.

Furthermore, we explore the relationship between the intrinsic matrix and the projection matrix based on the camera's field of view ($FoV$). First, we assume that the viewing volume is symmetric, which allows us to derive the following:

\begin{equation}
    \left\{ \begin{array}{c}
         r +l = 0  \\
         t +b = 0 \\
    \end{array}\right.  \left\{ \begin{array}{c}
         r = tan (FoV_{x}/2)\cdot n \\
         t = tan (FoV_{y}/2)\cdot n\\
    \end{array}\right.
\end{equation}
here $tan (FoV_{x}/2)$ and $tan (FoV_{y}) $ are $\frac{W}{2f_x}$ and $\frac{H}{2f_y}$, respectively. Additionally, $f_x$ and $f_y$ are the focal length parameters of the intrinsic matrix $\mathbf{K}$, given by: $\mathbf{K}= \left[ \begin{array}{ccc}
     f_x& 0 & c_x  \\
     0 &f_x & c_y \\
     0 & 0 & 1
\end{array} \right]$.


Therefore the Equation~\ref{eq:clip_C} can be represented. 
We express the transformation from camera coordinates to normalized clip space $\mathcal{P}$~\cite{zwicker2002ewa,ye2024gsplat},
\begin{equation}
    \mathcal{P} = \left[ \begin{array}{cccc}
        \frac{2f_x}{W} & 0  & 0  & 0\\
        0 & \frac{2f_y}{H}  & 0  & 0 \\
        0 & 0 & \frac{-(f+n)}{f-n} & \frac{-2fn}{f-n} \\
        0 & 0 & -1 & 0
    \end{array} \right]
\end{equation}
here, $W$ and $H$ represent the width and height of the image.


After obtaining the positions in the NDC cube, the screen coordinates are computed by applying a viewport transformation from the normalized device coordinates, which maps them into rendering pixel coordinates:

\begin{equation}
    \bm{\mu}_I = \left[\begin{array}{c}
          u \\
          v
    \end{array}\right] = \left[\begin{array}{c}
          \frac{W}{2}(x_{ndc}+1)+c_x \\
          \frac{H}{2}(y_{ndc}+1)+c_y
    \end{array}\right].
    \label{eq:intr_mu}
\end{equation}

Therefore, the Equation~\ref{eq:intr_mu} can be substituted as the mapping function $ \left[\begin{array}{c}
          u \\
          v
    \end{array}\right] = \phi(\bm{\mu}_C)$, and the function $\phi(\mathbf{x})$ for a position $\mathbf{x}=[x\;y\;z]^T$ can be denoted as: 
    
\begin{equation}
    \begin{split}
         \phi(\mathbf{x}) 
         & = \left[\begin{array}{ccc}
          \frac{W}{2} &0 &\frac{W}{2}+c_x \\
          0 & \frac{H}{2} & \frac{H}{2}+c_y \\
    \end{array}\right] \left[\begin{array}{c}
          x_{ndc}\\
          y_{ndc}\\
          1
    \end{array}\right] \\
    &  = \left[\begin{array}{ccc}
          \frac{W}{2} &0 &\frac{W}{2}+c_x \\
          0 & \frac{H}{2} & \frac{H}{2}+c_y \\
    \end{array}\right] \left[\begin{array}{c}
          x_{clip}/w_{clip}\\
          y_{ndc}/w_{clip}\\
          1
    \end{array}\right] \\
    &  = \left[\begin{array}{ccc}
          \frac{W}{2} &0 &\frac{W}{2}+c_x \\
          0 & \frac{H}{2} & \frac{H}{2}+c_y \\
    \end{array}\right] \left[\begin{array}{c}
          \frac{2f_x}{W}x /z\\
          \frac{2f_y}{H}y /z \\
          1
    \end{array}\right].
    \end{split}
\end{equation}

\subsection{Affine Approximation of Projective Matrix }
\label{subsec:affine_approx}

As is well-known, Gaussians are closed~\cite{zwicker2002ewa} under affine mappings. However, the projection transformations discussed are not affine, and the perspective projection of a 3D Gaussian does not directly yield a 2D Gaussian~\cite{huang2024error,li2025geogaussian}. 

To address this, we establish a local affine approximation \(\phi_k(\mathbf{x})\) of the projective transformation, defined by the first two terms of the Taylor expansion of \(\phi\) at the point \(\bm{\mu}\):

\begin{equation}
    \phi(\mathbf{x}) \approx \phi(\bm{\mu}) + \mathbf{J}(\mathbf{x} - \bm{\mu})
    \label{eq:taylor_expansion}
\end{equation}
where \(\bm{\mu}\) represents the center of the Gaussian in camera coordinates. The Jacobian \(\mathbf{J}\) is given by the partial derivatives of \(\phi\) evaluated at \(\bm{\mu}\),
\( 
\mathbf{J} = \frac{\partial \phi}{\partial \mathbf{x}} \Big|_{\mathbf{x} = \bm{\mu}}.
\).




Therefore, to approximate the projection of the covariance matrix \(\Sigma\) into pixel space, we use a first-order Taylor expansion around the point \(\bm{\mu}\) in the camera frame. Specifically, we compute the affine transformation \(\mathbf{J} \in \mathbb{R}^{2 \times 3}\) as follows:

\begin{equation}
    \mathbf{J} = \left[ \begin{array}{ccc}
         f_x/z_C & 0 & -f_xx_C/z_C^2 \\
         0 & f_y/z_C & -f_yy_C/z_C^2
    \end{array}\right].
    \label{eq:intr_j}
\end{equation}

In the world space, the 3D Gaussian function $\mathcal{G}$ is defined as:
\begin{equation}
    \mathcal{G}_{\bm{\Sigma}} (\mathbf{x} - \bm{\mu}) = \exp(\frac{-1}{2}(\mathbf{x} - \bm{\mu})^{T} \bm{\Sigma}^{-1} (\mathbf{x} - \bm{\mu}))
\end{equation}
where $\bm{\Sigma}$ and $\bm{\mu}$ are parameters for the Gaussian distribution.

When we transform the 3D Gaussian from the world coordinates to the camera coordinates via the camera pose $\mathbf{T}= \left[ \begin{array}{cc}
    \mathbf{W} & \mathbf{t}  \\
    0 & 1 
\end{array}\right] \in SE(3) $, the transformed Gaussian can be represented as $\mathcal{G}_{\mathbf{W}\bm{\Sigma}\mathbf{W}^T} (\mathbf{W}\mathbf{x} - \mathbf{W}\bm{\mu})$.


When we project the Gaussian function into screen space using the affine approximation defined in Equation~\ref{eq:taylor_expansion}, the covariance matrix of the 2D Gaussian can be expressed as \(\mathbf{J} \mathbf{W} \bm{\Sigma} \mathbf{W}^T \mathbf{J}^T\).  
The optimization for the intrinsic matrix can then be derived based on the relationships represented in Equations~\ref{eq:intr_mu} and~\ref{eq:intr_j}.

\begin{figure*}
    \centering
    \includegraphics[width=\textwidth]{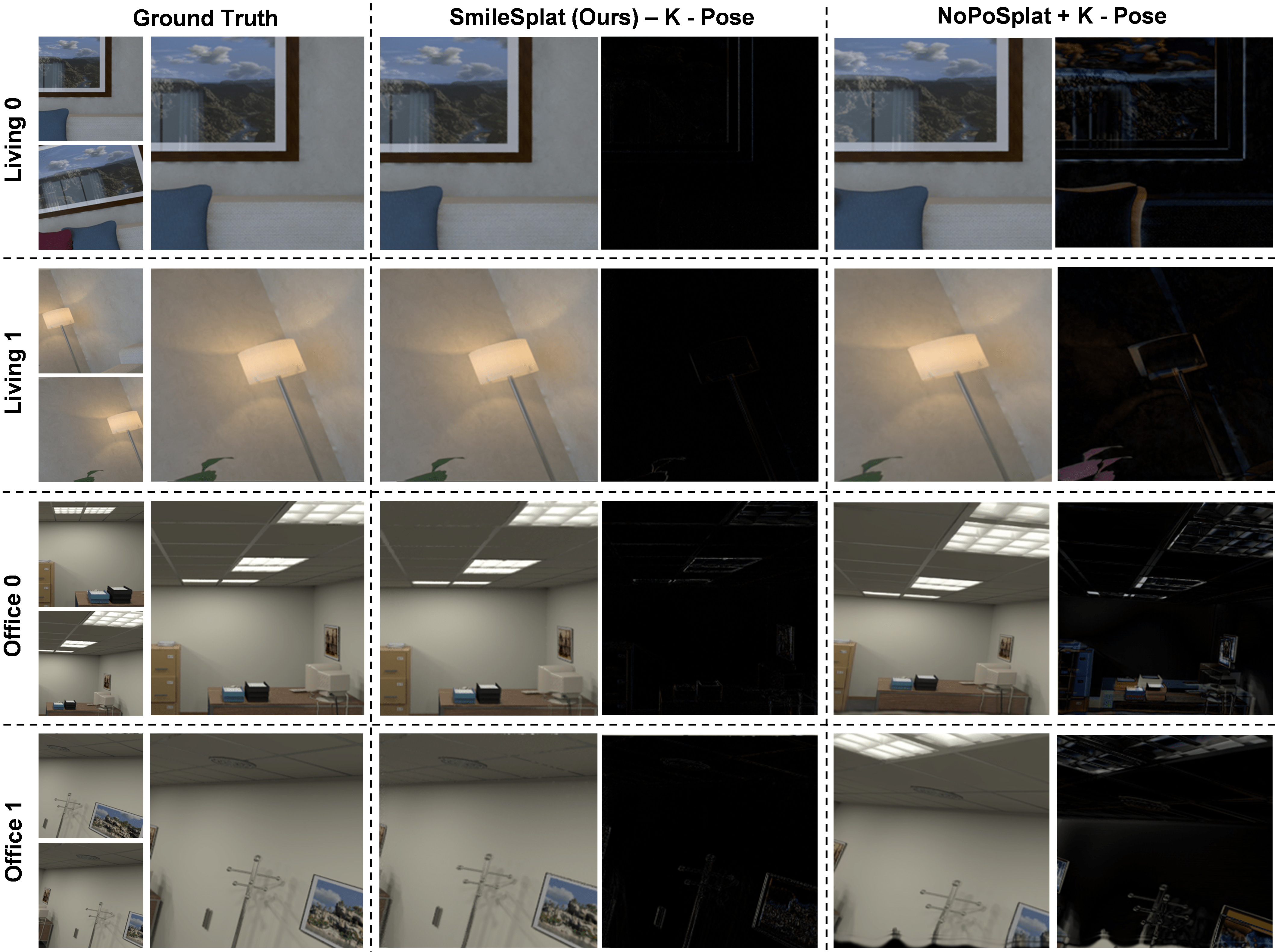}
    \caption{
    Comparisons of novel view rendering on the ICL-NUIM~\cite{handa2014benchmark} dataset are presented. The rendering results include the RGB images in the left column and the differences from the ground truth in the right column. $\pm$ K and $\pm$ Pose indicate whether the intrinsics and extrinsics are required or not, respectively.}
    \label{fig:cross-dataset-ICL}
\end{figure*}

\begin{figure*}
    \centering
    \includegraphics[width=\textwidth]{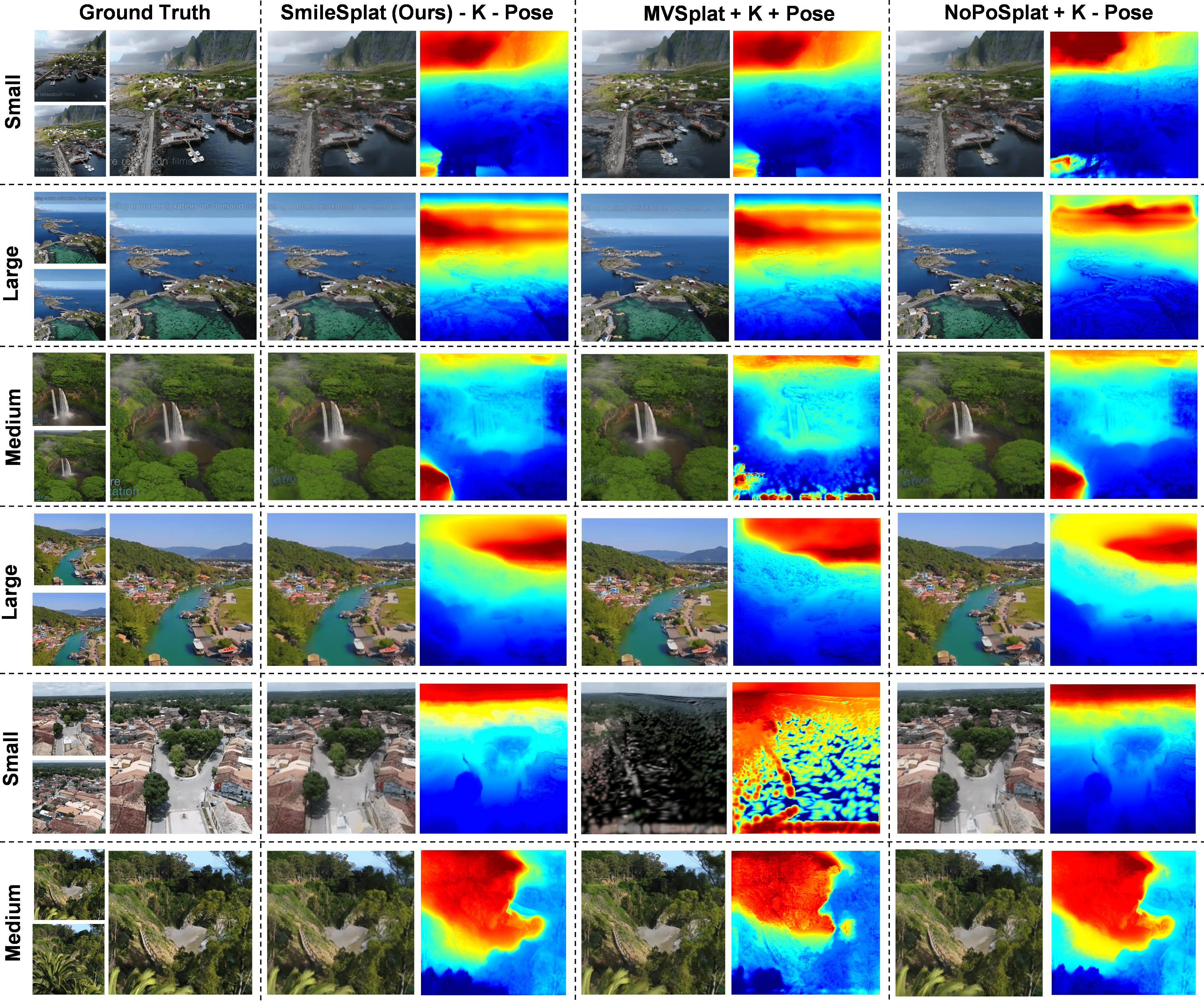}
    \caption{
    Comparisons of novel view rendering on the ACID~\cite{liu2021infinite} dataset are presented. The rendering results include the RGB images in the left column and the differences from the ground truth in the right column. $\pm$ K and $\pm$ Pose indicate whether the intrinsics and extrinsics are required or not, respectively.}
    \label{fig:render_ACID}
\end{figure*}

\begin{figure*}
    \centering
    \begin{subfigure}[b]{\textwidth}
    \resizebox{0.034\linewidth}{!}{
     	\begin{tikzpicture}
        \node[rotate=270,] at (0, 0) {};
        \node[rotate=270,] at (0, 0.5) {\tiny GT V1};
        \node[rotate=270,] at (0, 1.75) {\tiny Rendered V1};
        \node[rotate=270,] at (0, -0.7) {\tiny Difference};
        \node[rotate=270,] at (0, -0.8) {};
    	\end{tikzpicture}}
    \includegraphics[width=0.09\textwidth]{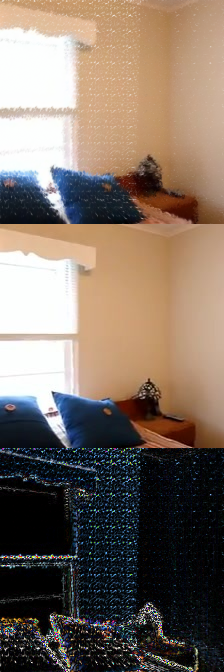}
    \includegraphics[width=0.09\textwidth]{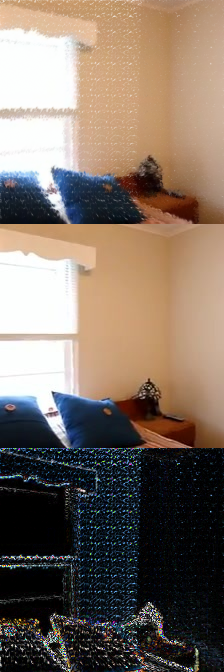}
    \includegraphics[width=0.09\textwidth]{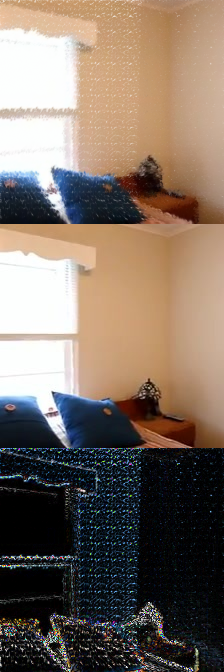}
    \includegraphics[width=0.09\textwidth]{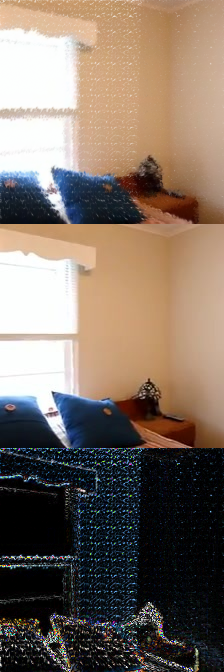}
    \includegraphics[width=0.09\textwidth]{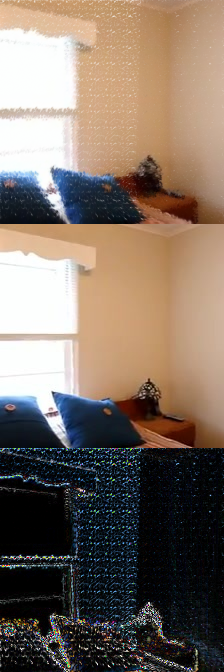}
    \includegraphics[width=0.09\textwidth]{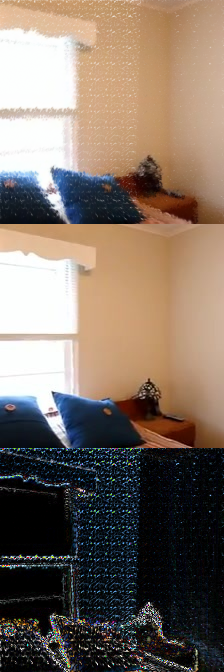}
    \includegraphics[width=0.09\textwidth]{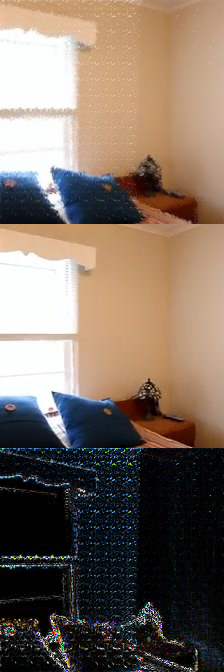}
    \includegraphics[width=0.09\textwidth]{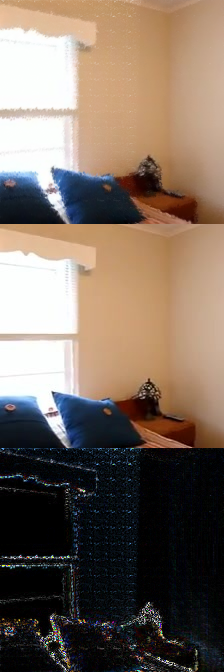}
    \includegraphics[width=0.09\textwidth]{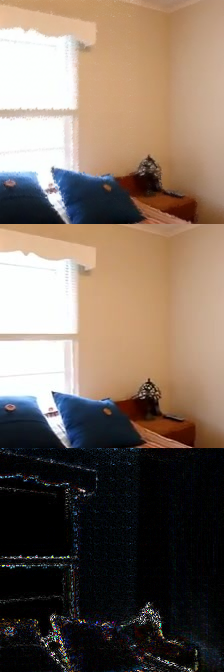}
    \includegraphics[width=0.09\textwidth]{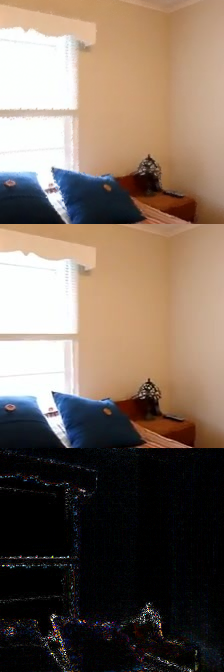}
     \begin{tikzpicture}
        \node[] at (0, 0) {};
        \node[] at (-7.5, 0) {};
        \node[] at (-6.8+0.6, 0) {I=0};
        \node[] at (-6.8+2.35, 0) {I=2};
        \node[] at (-6.8+0.6+1.75*2, 0) {I=4};
        \node[] at (-6.8+0.55+1.75*3, 0) {I=6};
        \node[] at (-6.8+0.48+1.75*4, 0) {I=8};
        \node[] at (-6.8+0.4+1.75*5, 0) {I=10};
        \node[] at (-6.8+0.38+1.75*6, 0) {I=12};
        \node[] at (-6.8+0.32+1.75*7, 0) {I=14};
        \node[] at (-6.8+0.22+1.75*8, 0) {I=16};
        \node[] at (-6.8+0.18+1.75*9, 0) {I=18};
    \end{tikzpicture}\\
    \resizebox{0.034\linewidth}{!}{
     	\begin{tikzpicture}
        \node[rotate=270,] at (0, 0) {};
        \node[rotate=270,] at (0, 0.5) {\tiny GT V2};
        \node[rotate=270,] at (0, 1.75) {\tiny Rendered V2};
        \node[rotate=270,] at (0, -0.7) {\tiny Difference};
        \node[rotate=270,] at (0, -0.8) {};
    	\end{tikzpicture}}
    \includegraphics[width=0.09\textwidth]{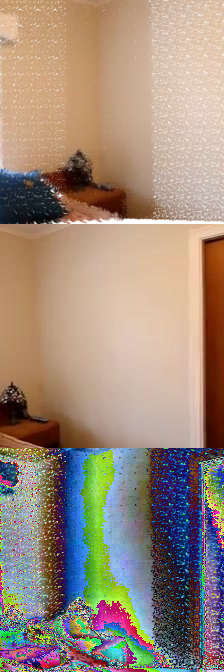}
    \includegraphics[width=0.09\textwidth]{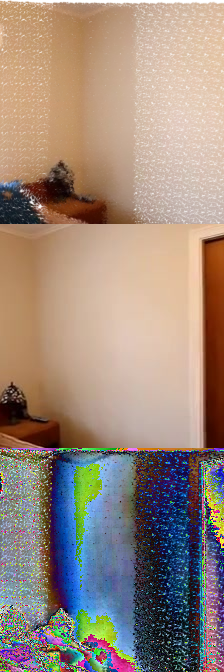}
    \includegraphics[width=0.09\textwidth]{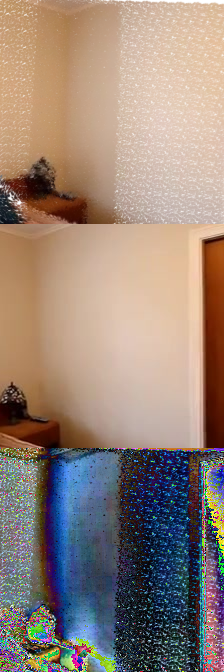}
    \includegraphics[width=0.09\textwidth]{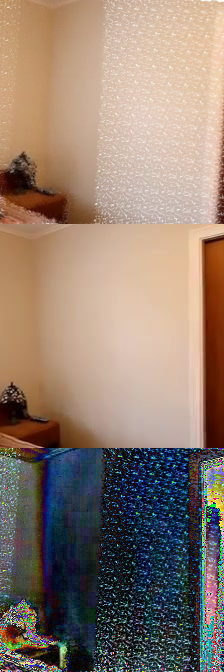}
    \includegraphics[width=0.09\textwidth]{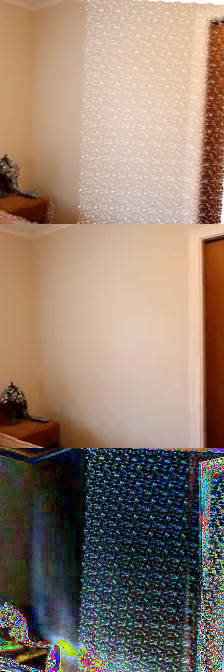}
    \includegraphics[width=0.09\textwidth]{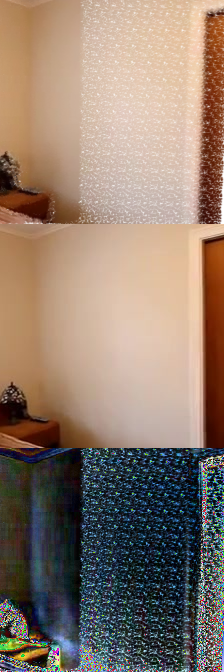}
    \includegraphics[width=0.09\textwidth]{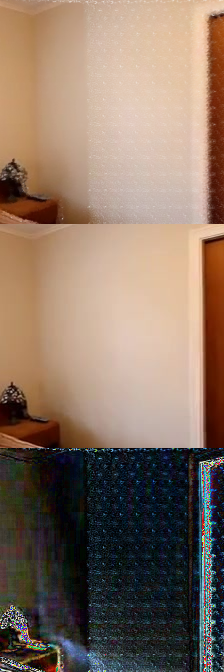}
    \includegraphics[width=0.09\textwidth]{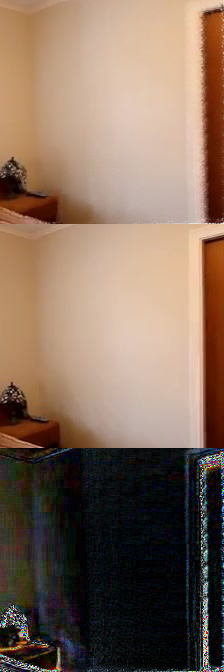}
    \includegraphics[width=0.09\textwidth]{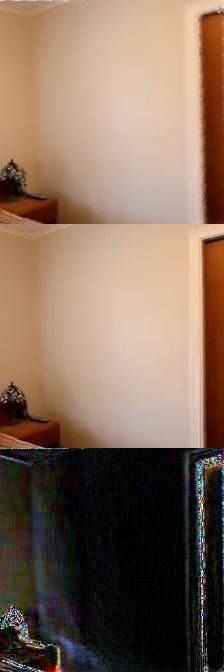}
    \includegraphics[width=0.09\textwidth]{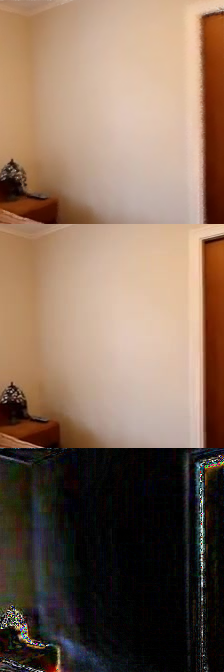} 
    \begin{tikzpicture}
        \node[] at (0, 0) {};
        \node[] at (-7.5, 0) {};
        \node[] at (-6.8+0.6, 0) {I=0};
        \node[] at (-6.8+2.35, 0) {I=4};
        \node[] at (-6.8+0.6+1.75*2, 0) {I=8};
        \node[] at (-6.8+0.55+1.75*3, 0) {I=12};
        \node[] at (-6.8+0.48+1.75*4, 0) {I=16};
        \node[] at (-6.8+0.4+1.75*5, 0) {I=20};
        \node[] at (-6.8+0.38+1.75*6, 0) {I=24};
        \node[] at (-6.8+0.32+1.75*7, 0) {I=28};
        \node[] at (-6.8+0.22+1.75*8, 0) {I=32};
        \node[] at (-6.8+0.18+1.75*9, 0) {I=36};
    \end{tikzpicture}\\
     \resizebox{0.034\linewidth}{!}{
     	\begin{tikzpicture}
        \node[rotate=270,] at (0, 0) {};
        \node[rotate=270,] at (0, 0.5) {\tiny GT View};
        \node[rotate=270,] at (0, 1.75) {\tiny Novel View};
        \node[rotate=270,] at (0, -0.7) {\tiny Difference};
        \node[rotate=270,] at (0, -0.8) {};
    	\end{tikzpicture}}
    \includegraphics[width=0.09\textwidth]{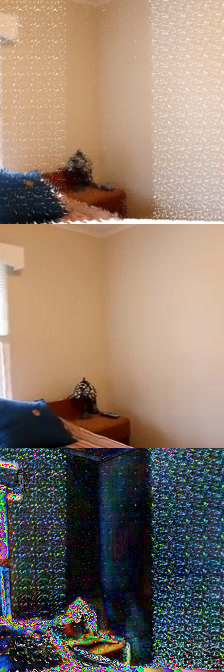}
    \includegraphics[width=0.09\textwidth]{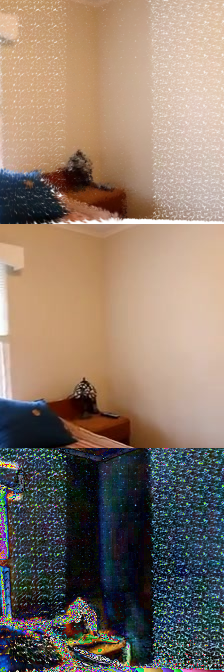}
    \includegraphics[width=0.09\textwidth]{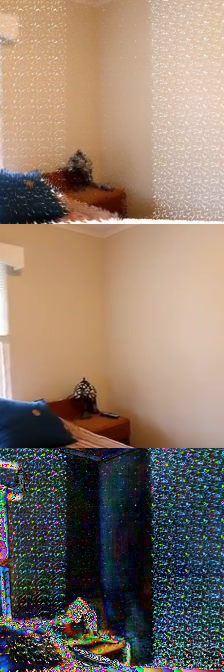}
    \includegraphics[width=0.09\textwidth]{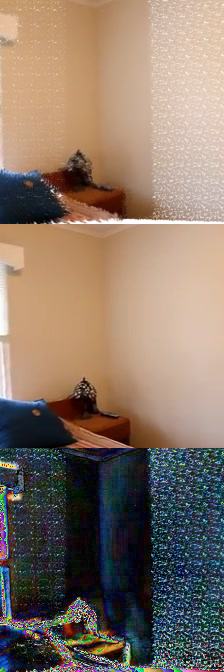}
    \includegraphics[width=0.09\textwidth]{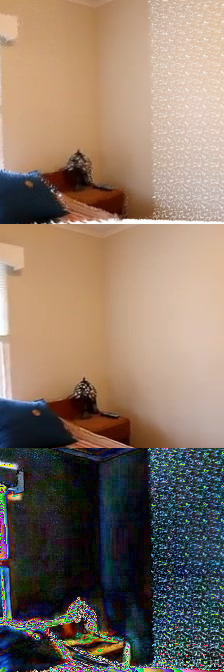}
    \includegraphics[width=0.09\textwidth]{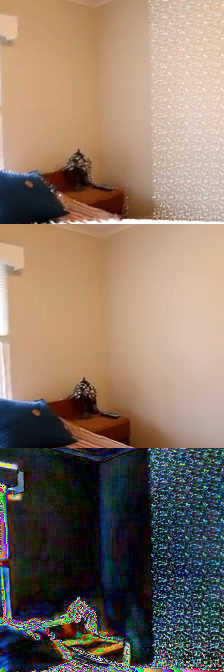}
    \includegraphics[width=0.09\textwidth]{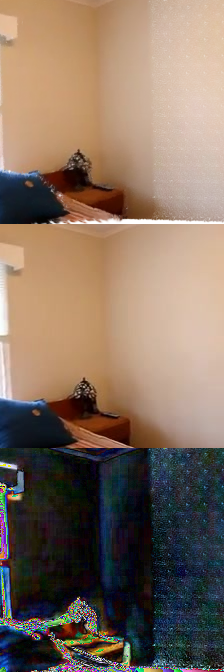}
    \includegraphics[width=0.09\textwidth]{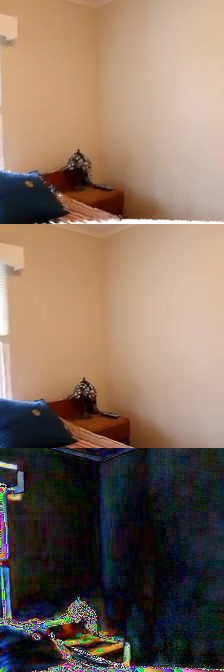}
    \includegraphics[width=0.09\textwidth]{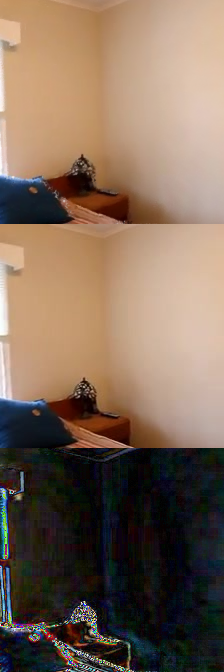}
    \includegraphics[width=0.09\textwidth]{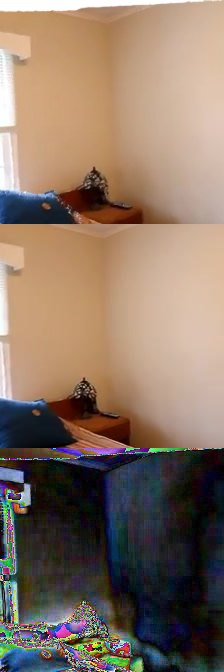} 
    \begin{tikzpicture}
        \node[] at (0, 0) {};
        \node[] at (-7.5, 0) {};
        \node[] at (-6.8+0.6, 0) {I=0};
        \node[] at (-6.8+2.35, 0) {I=4};
        \node[] at (-6.8+0.6+1.75*2, 0) {I=8};
        \node[] at (-6.8+0.55+1.75*3, 0) {I=12};
        \node[] at (-6.8+0.48+1.75*4, 0) {I=16};
        \node[] at (-6.8+0.4+1.75*5, 0) {I=20};
        \node[] at (-6.8+0.38+1.75*6, 0) {I=24};
        \node[] at (-6.8+0.32+1.75*7, 0) {I=28};
        \node[] at (-6.8+0.22+1.75*8, 0) {I=32};
        \node[] at (-6.8+0.18+1.75*9, 0) {I=36};
    \end{tikzpicture}\\
    \subcaption{Optimization (the first 36 iterations) for intrinsic and extrinsic parameters.}    
    \end{subfigure}
    \begin{subfigure}[b]{0.40\textwidth}
    \centering
        \includegraphics[width=0.24\textwidth]{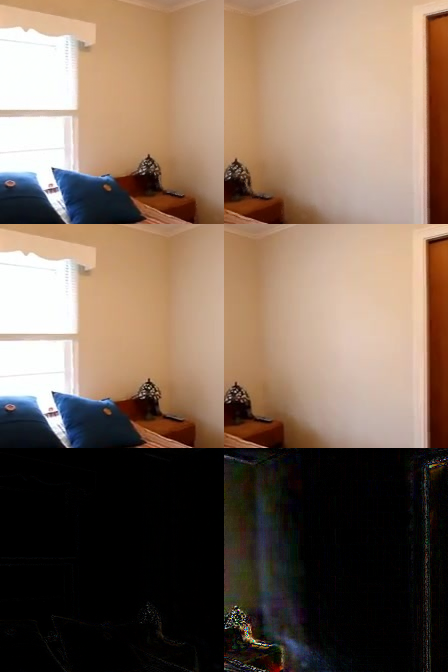}
        \includegraphics[width=0.24\textwidth]{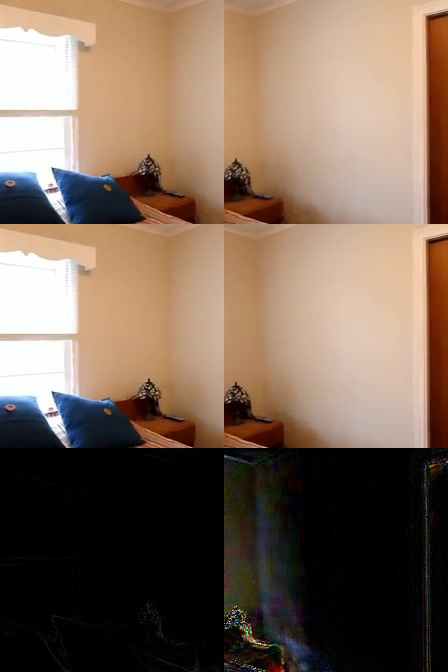}
        \includegraphics[width=0.24\textwidth]{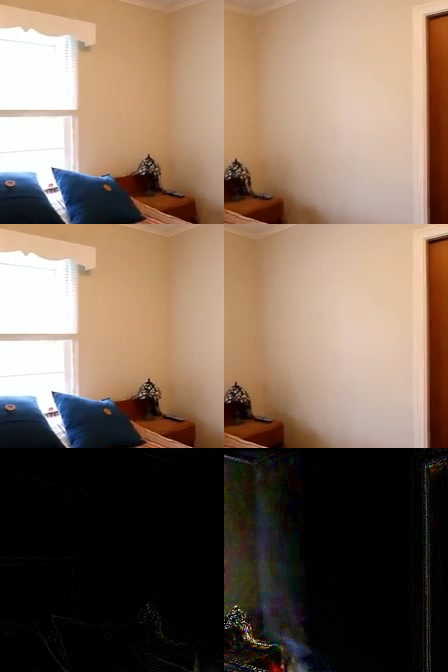}
        \includegraphics[width=0.24\textwidth]{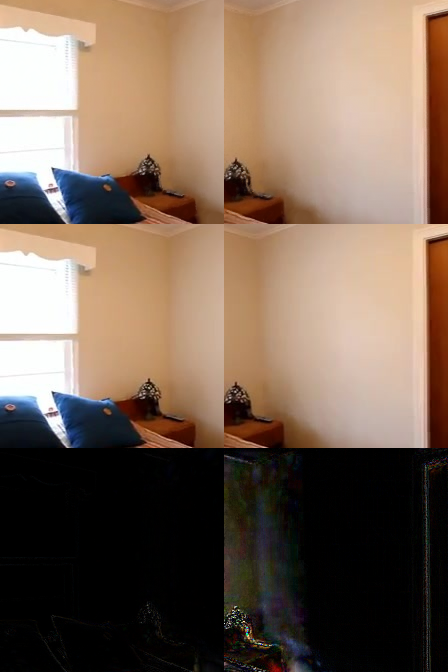}
        \begin{tikzpicture}
        \node[] at (0, 0) {};
        \node[] at (-7.5, 0) {};
        \node[] at (-6.8, 0) {I=60};
        \node[] at (-6.8+1.75, 0) {I=70};
        \node[] at (-6.8+1.75*2, 0) {I=80};
        \node[] at (-6.8+1.75*3, 0) {I=90};
        \node[] at (2, 0) {};
    \end{tikzpicture}\\
        \subcaption{Bundle adjusting Gaussian Splatting for two views.}
    \end{subfigure}
    \begin{subfigure}[b]{0.57\textwidth}
     \centering
        \includegraphics[width=0.49\linewidth]{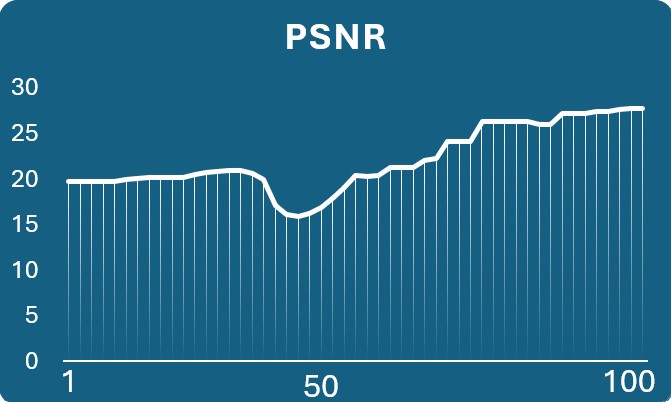}
        \includegraphics[width=0.49\linewidth]{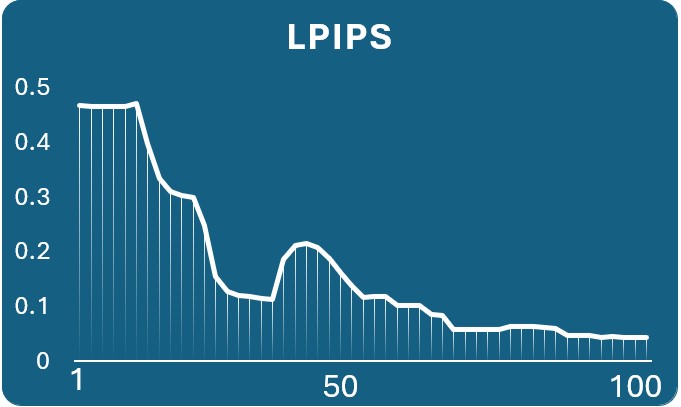}
        \caption{The curves of PSNR and LPIPS of novel view rendering during the refinement process}
    \end{subfigure}
    \caption{The entire optimization process. To clearly visualize the difference maps, we have magnified the error values by a \textbf{factor of 5}.}
    \label{fig:opti_50_36}
\end{figure*}
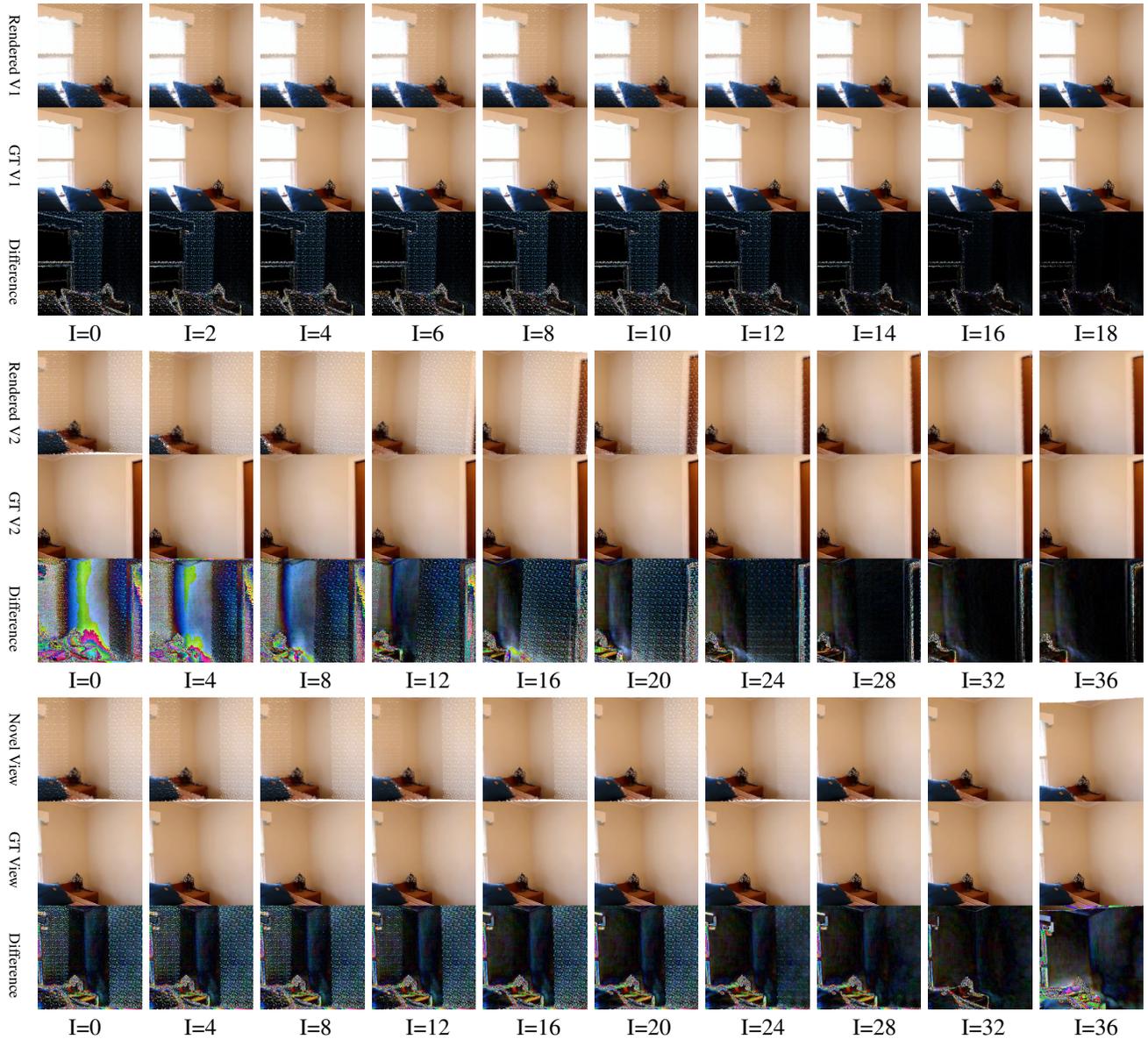
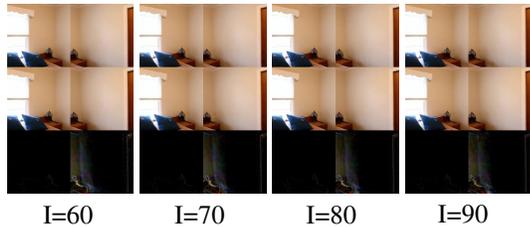
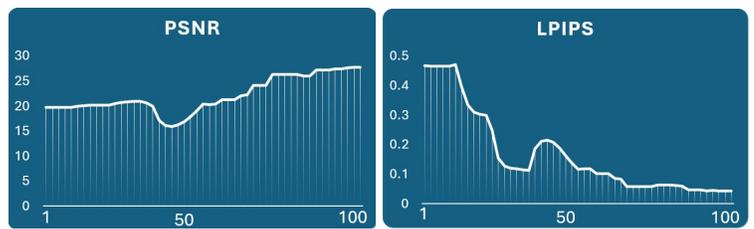

\begin{figure*}
    \centering
    \resizebox{0.034\linewidth}{!}{
     	\begin{tikzpicture}
        \node[rotate=270,] at (0, 0) {};
        \node[rotate=270,] at (0, 0.5) {\tiny GT V1};
        \node[rotate=270,] at (0, 1.75) {\tiny Rendered V1};
        \node[rotate=270,] at (0, -0.7) {\tiny Difference};
        \node[rotate=270,] at (0, -0.8) {};
    	\end{tikzpicture}}
    \includegraphics[width=0.09\textwidth]{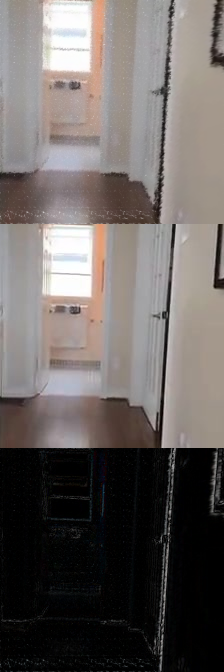}
    \includegraphics[width=0.09\textwidth]{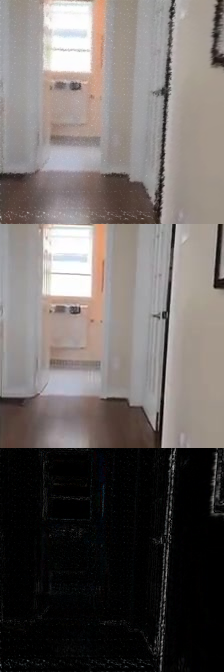}
    \includegraphics[width=0.09\textwidth]{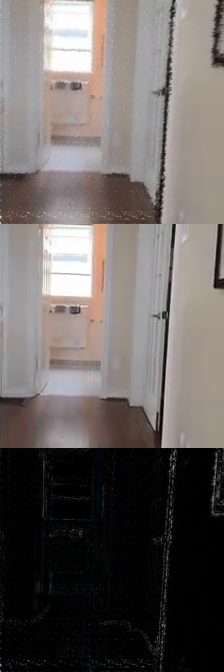}
    \includegraphics[width=0.09\textwidth]{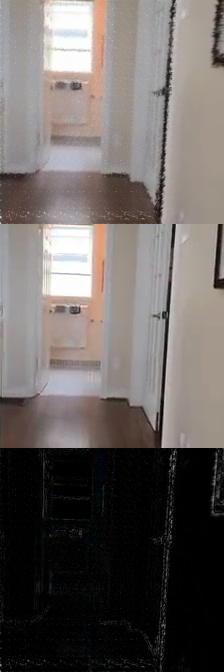}
    \includegraphics[width=0.09\textwidth]{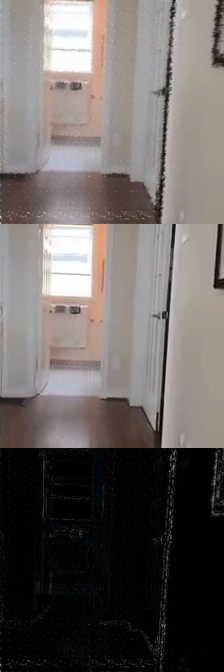}
    \includegraphics[width=0.09\textwidth]{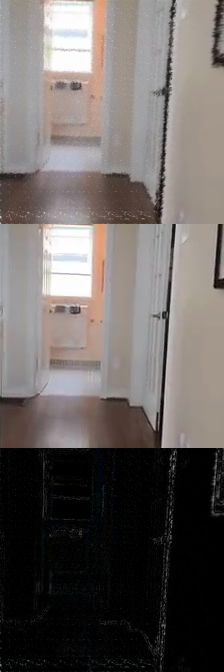}
    \includegraphics[width=0.09\textwidth]{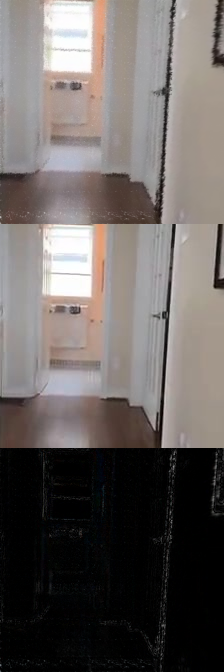}
    \includegraphics[width=0.09\textwidth]{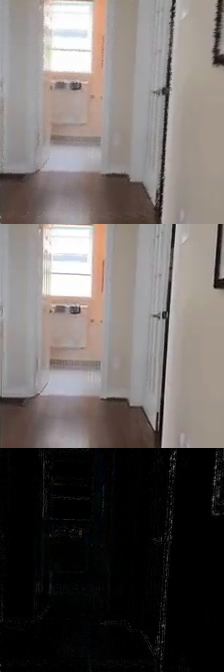}
    \includegraphics[width=0.09\textwidth]{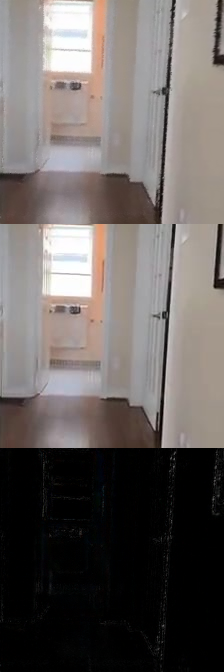}
    \includegraphics[width=0.09\textwidth]{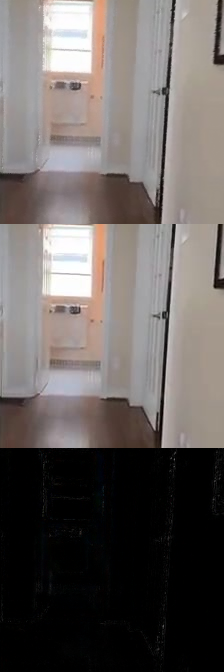}\\
     \begin{tikzpicture}
        \node[] at (0, 0) {};
        \node[] at (-7.5, 0) {};
        \node[] at (-6.8+0.1, 0) {I=0};
        \node[] at (-6.8+1.7, 0) {I=2};
        \node[] at (-6.8+0.18+1.6*2, 0) {I=4};
        \node[] at (-6.8+0.2+1.6*3, 0) {I=6};
        \node[] at (-6.8+0.3+1.6*4, 0) {I=8};
        \node[] at (-6.8+0.43+1.6*5, 0) {I=10};
        \node[] at (-6.8+0.5+1.6*6, 0) {I=12};
        \node[] at (-6.8+0.53+1.6*7, 0) {I=14};
        \node[] at (-6.8+0.57+1.6*8, 0) {I=16};
        \node[] at (-6.8+0.61+1.6*9, 0) {I=18};
    \end{tikzpicture}\\
    \resizebox{0.034\linewidth}{!}{
     	\begin{tikzpicture}
        \node[rotate=270,] at (0, 0) {};
        \node[rotate=270,] at (0, 0.5) {\tiny GT V2};
        \node[rotate=270,] at (0, 1.75) {\tiny Rendered V2};
        \node[rotate=270,] at (0, -0.7) {\tiny Difference};
        \node[rotate=270,] at (0, -0.8) {};
    	\end{tikzpicture}}
    \includegraphics[width=0.09\textwidth]{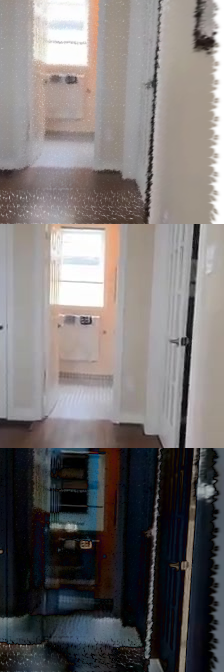}
    \includegraphics[width=0.09\textwidth]{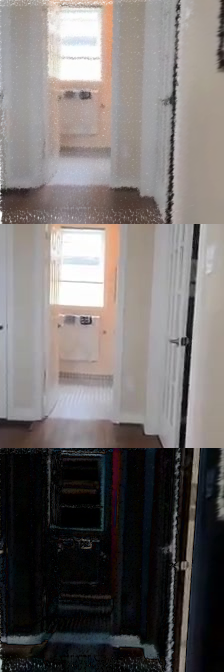}
    \includegraphics[width=0.09\textwidth]{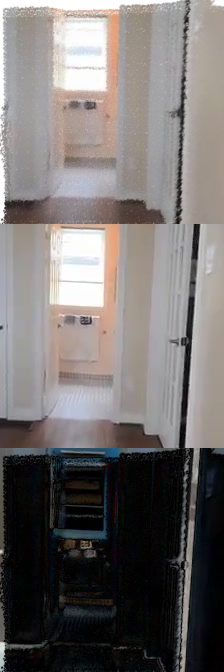}
    \includegraphics[width=0.09\textwidth]{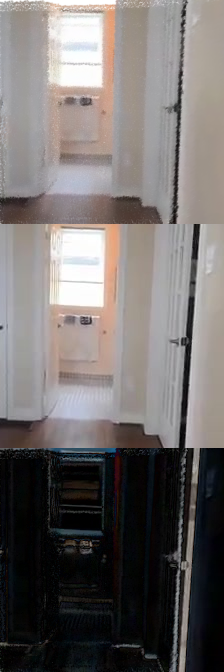}
    \includegraphics[width=0.09\textwidth]{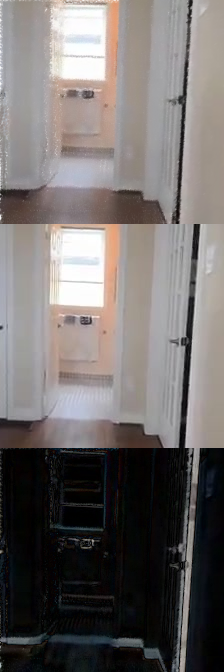}
    \includegraphics[width=0.09\textwidth]{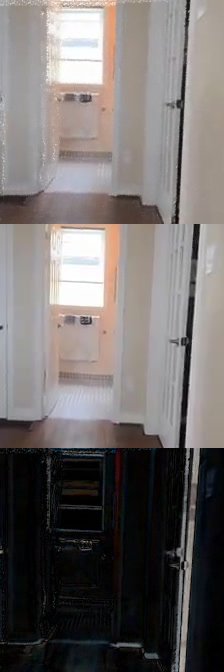}
    \includegraphics[width=0.09\textwidth]{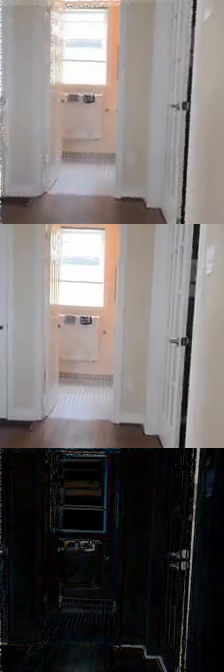}
    \includegraphics[width=0.09\textwidth]{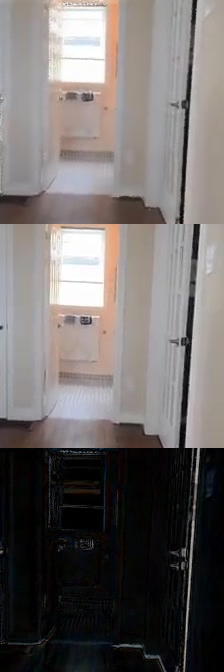}
    \includegraphics[width=0.09\textwidth]{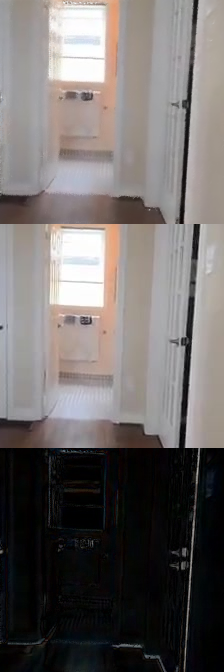}
    \includegraphics[width=0.09\textwidth]{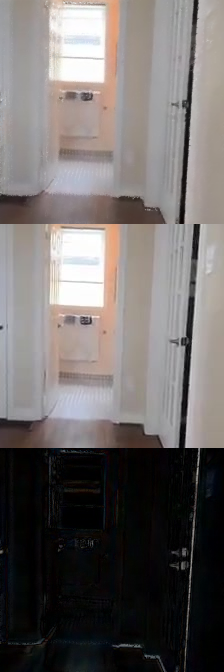} 
    \begin{tikzpicture}
        \node[] at (0, 0) {};
        \node[] at (-7.5, 0) {};
        \node[] at (-6.8+0.4, 0) {I=0};
        \node[] at (-6.8+2, 0) {I=4};
        \node[] at (-6.8+0.35+1.7*2, 0) {I=8};
        \node[] at (-6.8+0.35+1.7*3, 0) {I=12};
        \node[] at (-6.8+0.25+1.7*4, 0) {I=16};
        \node[] at (-6.8+0.25+1.7*5, 0) {I=20};
        \node[] at (-6.8+0.25+1.7*6, 0) {I=24};
        \node[] at (-6.8+0.2+1.7*7, 0) {I=28};
        \node[] at (-6.8+0.15+1.7*8, 0) {I=32};
        \node[] at (-6.8+0.10+1.7*9, 0) {I=36};
    \end{tikzpicture}\\
     \resizebox{0.034\linewidth}{!}{
     	\begin{tikzpicture}
        \node[rotate=270,] at (0, 0) {};
        \node[rotate=270,] at (0, 0.5) {\tiny GT View};
        \node[rotate=270,] at (0, 1.75) {\tiny Novel View};
        \node[rotate=270,] at (0, -0.7) {\tiny Difference};
        \node[rotate=270,] at (0, -0.8) {};
    	\end{tikzpicture}}
    \includegraphics[width=0.09\textwidth]{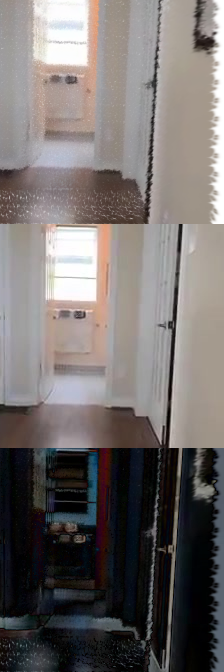}
    \includegraphics[width=0.09\textwidth]{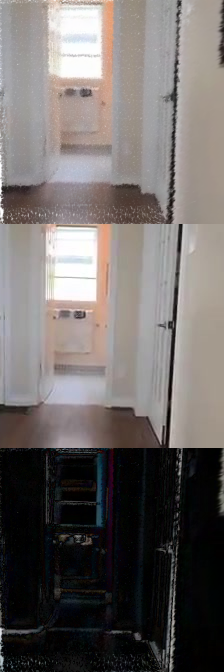}
    \includegraphics[width=0.09\textwidth]{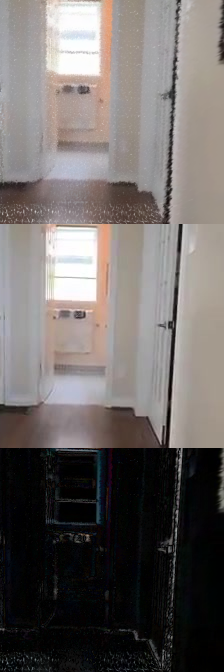}
    \includegraphics[width=0.09\textwidth]{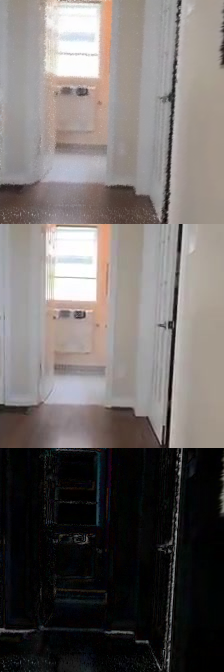}
    \includegraphics[width=0.09\textwidth]{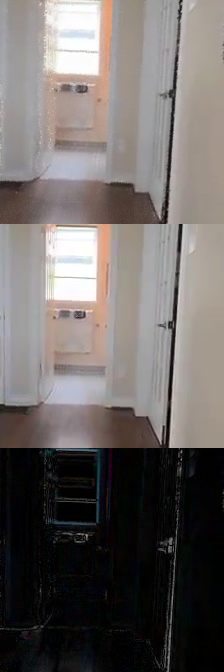}
    \includegraphics[width=0.09\textwidth]{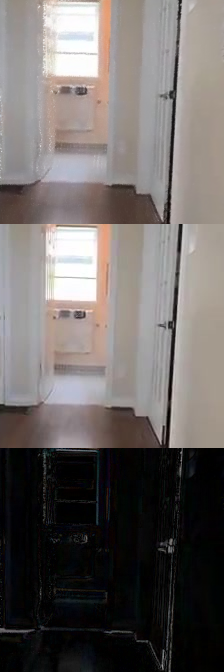}
    \includegraphics[width=0.09\textwidth]{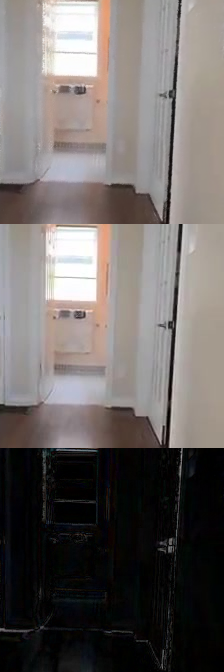}
    \includegraphics[width=0.09\textwidth]{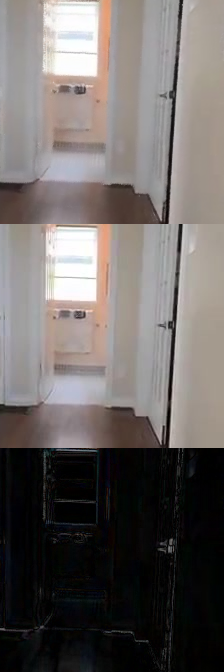}
    \includegraphics[width=0.09\textwidth]{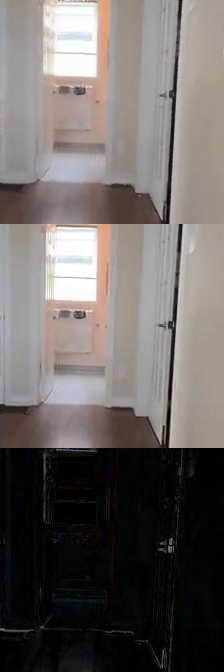}
    \includegraphics[width=0.09\textwidth]{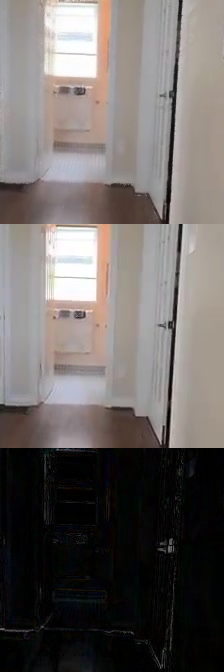} 
    \begin{tikzpicture}
        \node[] at (0, 0) {};
        \node[] at (-7.5, 0) {};
        \node[] at (-6.8+0.4, 0) {I=0};
        \node[] at (-6.8+2, 0) {I=4};
        \node[] at (-6.8+0.35+1.7*2, 0) {I=8};
        \node[] at (-6.8+0.35+1.7*3, 0) {I=12};
        \node[] at (-6.8+0.25+1.7*4, 0) {I=16};
        \node[] at (-6.8+0.25+1.7*5, 0) {I=20};
        \node[] at (-6.8+0.25+1.7*6, 0) {I=24};
        \node[] at (-6.8+0.2+1.7*7, 0) {I=28};
        \node[] at (-6.8+0.15+1.7*8, 0) {I=32};
        \node[] at (-6.8+0.10+1.7*9, 0) {I=36};
    \end{tikzpicture}\\
    \caption{Optimization process (the first 36 iterations) for intrinsic and extrinsic parameters.}
    \label{fig:opti_0_36}
\end{figure*}
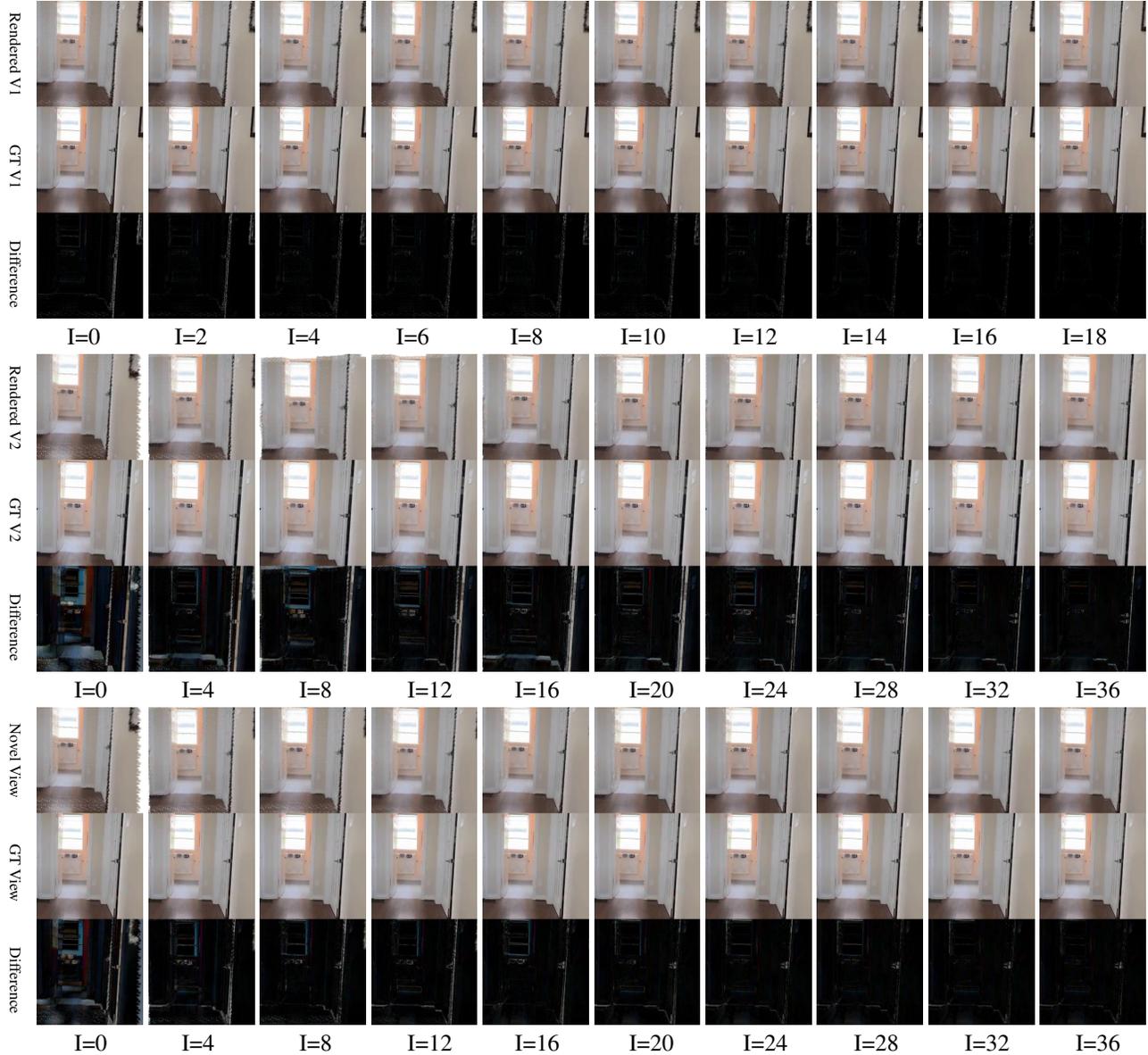



\begin{figure*}
    \centering
    \begin{subfigure}[b]{\textwidth}
     \resizebox{0.034\linewidth}{!}{
     	\begin{tikzpicture}
        \node[rotate=270,] at (0, 0) {};
        \node[rotate=270,] at (0, 0.7) {\tiny GT View };
        \node[rotate=270,] at (0, 2.0) {\tiny Novel View};
        \node[rotate=270,] at (0, -0.5) {\tiny Difference};
        \node[rotate=270,] at (0, -0.8) {};
    	\end{tikzpicture}}
    \includegraphics[width=0.09\textwidth]{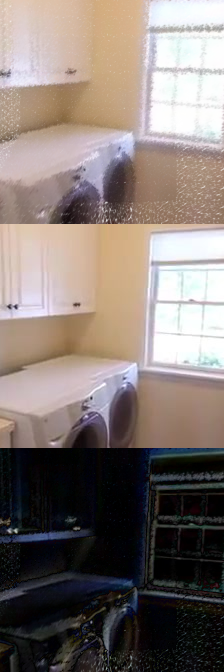}
    \includegraphics[width=0.09\textwidth]{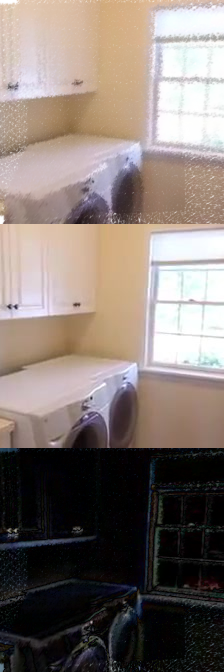}
    \includegraphics[width=0.09\textwidth]{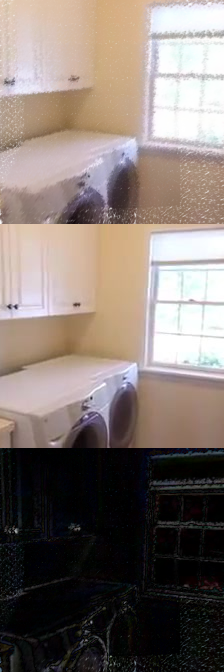}
    \includegraphics[width=0.09\textwidth]{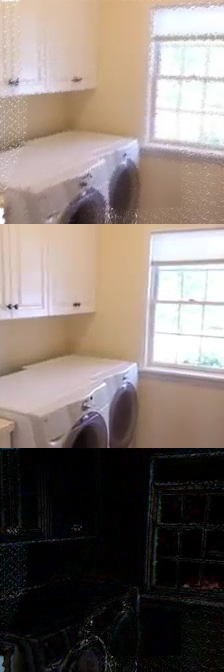}
    \includegraphics[width=0.09\textwidth]{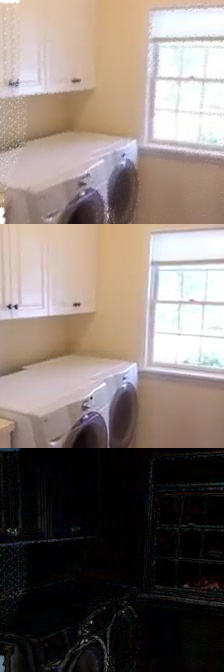}
    \includegraphics[width=0.09\textwidth]{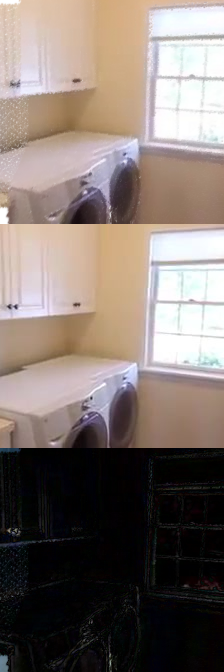}
    \includegraphics[width=0.09\textwidth]{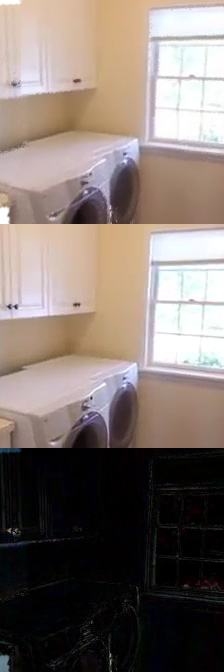}
    \includegraphics[width=0.09\textwidth]{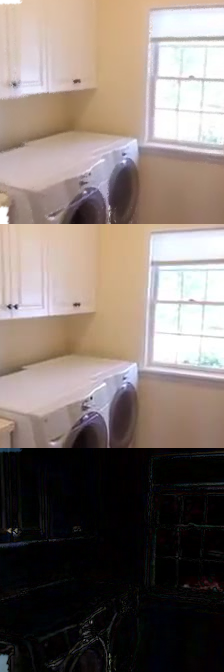}
    \includegraphics[width=0.09\textwidth]{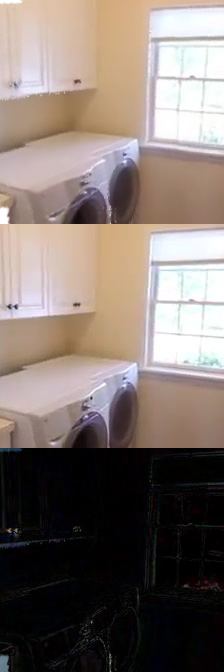}
    \includegraphics[width=0.09\textwidth]{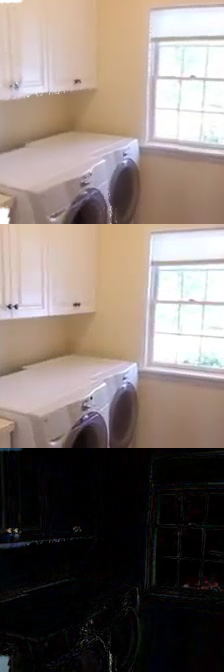} 
     \begin{tikzpicture}
        \node[] at (0, 0) {};
        \node[] at (-7.5, 0) {};
        \node[] at (-6.8+0.6, 0) {I=0};
        \node[] at (-6.8+2+0.3, 0) {I=4};
        \node[] at (-6.8+0.35+0.3+1.7*2, 0) {I=8};
        \node[] at (-6.8+0.35+0.35+1.7*3, 0) {I=12};
        \node[] at (-6.8+0.25+0.35+1.7*4, 0) {I=16};
        \node[] at (-6.8+0.25+0.3+1.7*5, 0) {I=20};
        \node[] at (-6.8+0.25+0.32+1.7*6, 0) {I=24};
        \node[] at (-6.8+0.2+0.35+1.7*7, 0) {I=28};
        \node[] at (-6.8+0.15+0.5+1.7*8, 0) {I=32};
        \node[] at (-6.8+0.10+0.5+1.7*9, 0) {I=36};
    \end{tikzpicture}
    \subcaption{First 36 iterations in refinement for View 1, View 2, and the novel viewpoint.}
    \end{subfigure}
    \begin{subfigure}[b]{0.3\textwidth}
    \includegraphics[width=0.98\linewidth, clip, trim={900 350 900 250}]{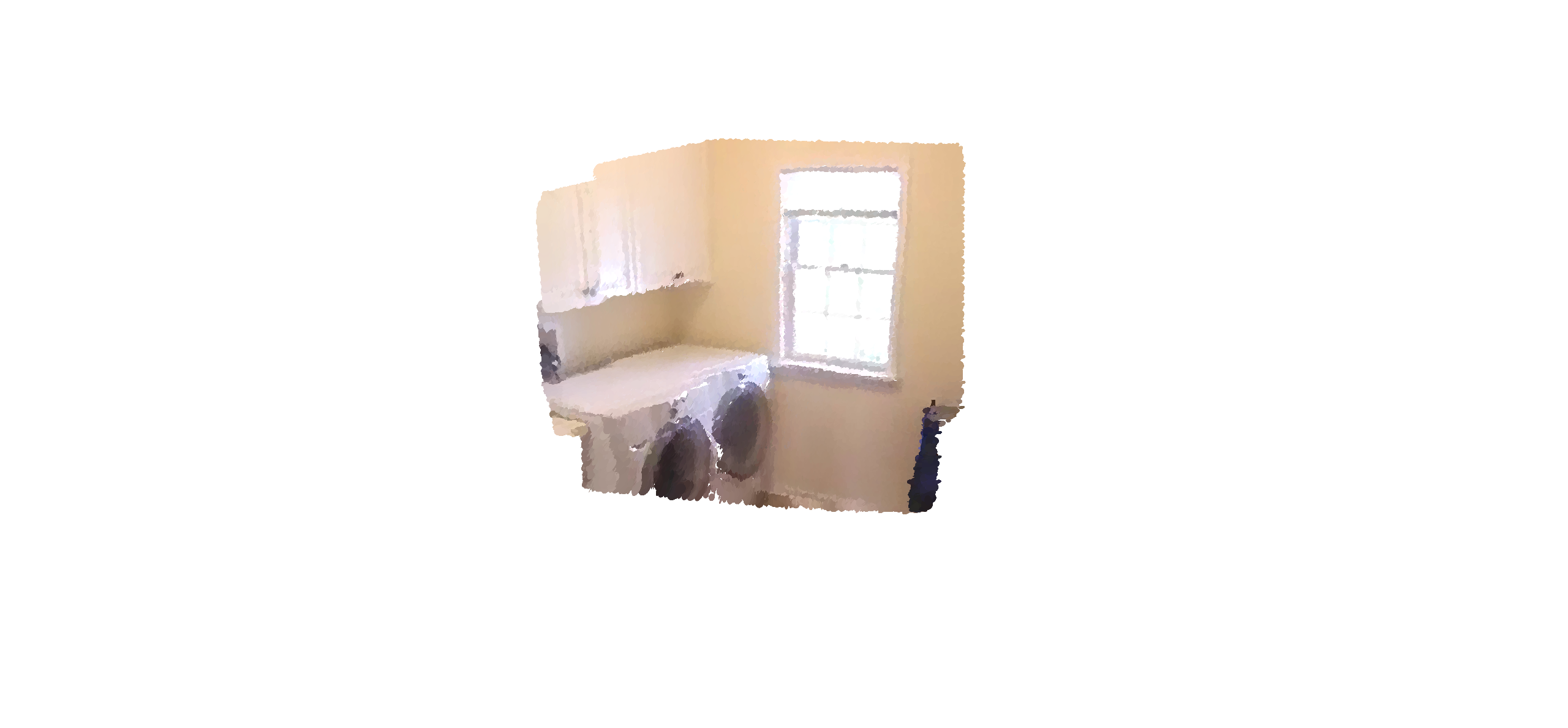}
    \subcaption{3D Gaussian map}
    \end{subfigure}
    \begin{subfigure}[b]{0.3\textwidth}
    \includegraphics[width=0.98\linewidth, clip, trim={900 350 900 250}]{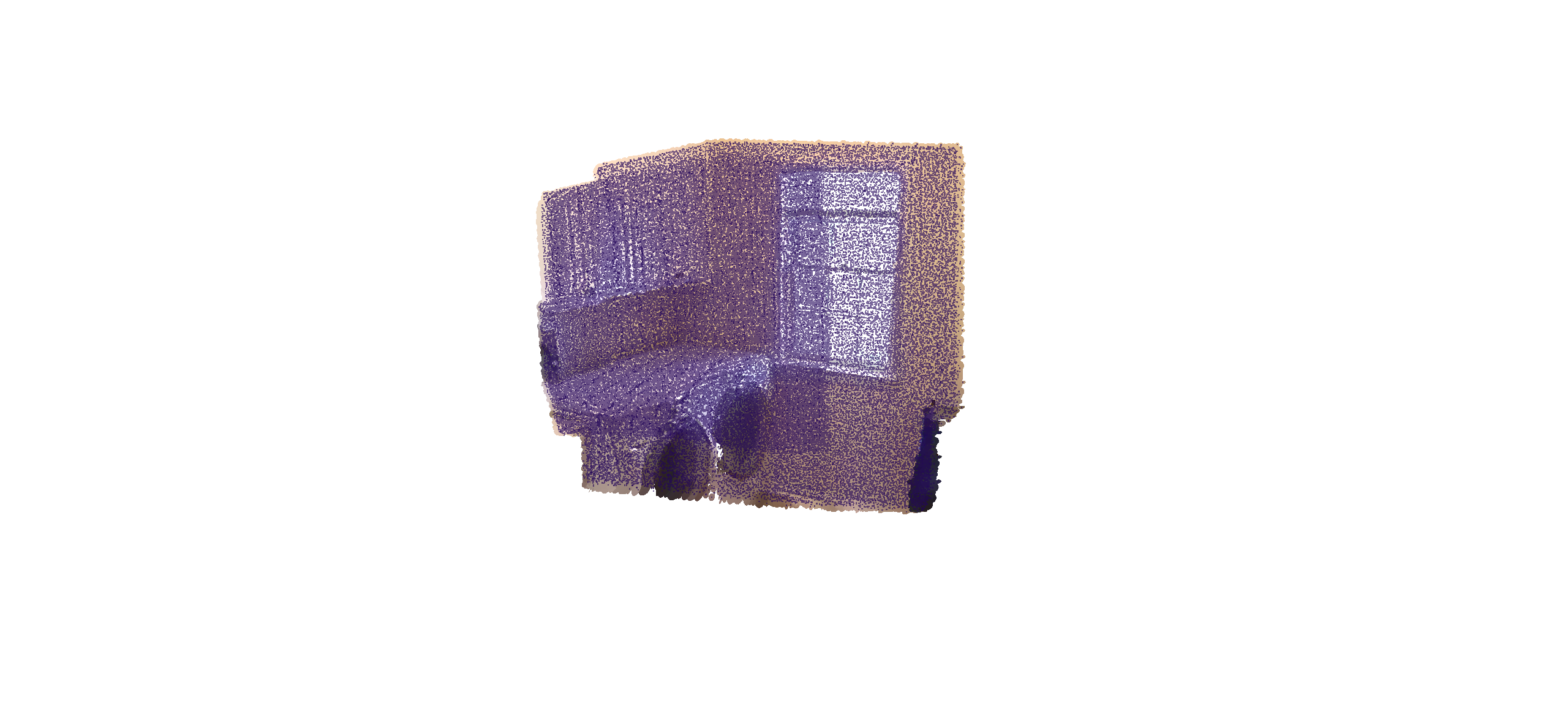}
    \subcaption{Visual overlaps}
    \end{subfigure}
    \begin{subfigure}[b]{0.3\textwidth}
    \centering
    \includegraphics[width=0.46\linewidth, clip, trim={900 350 900 250}]{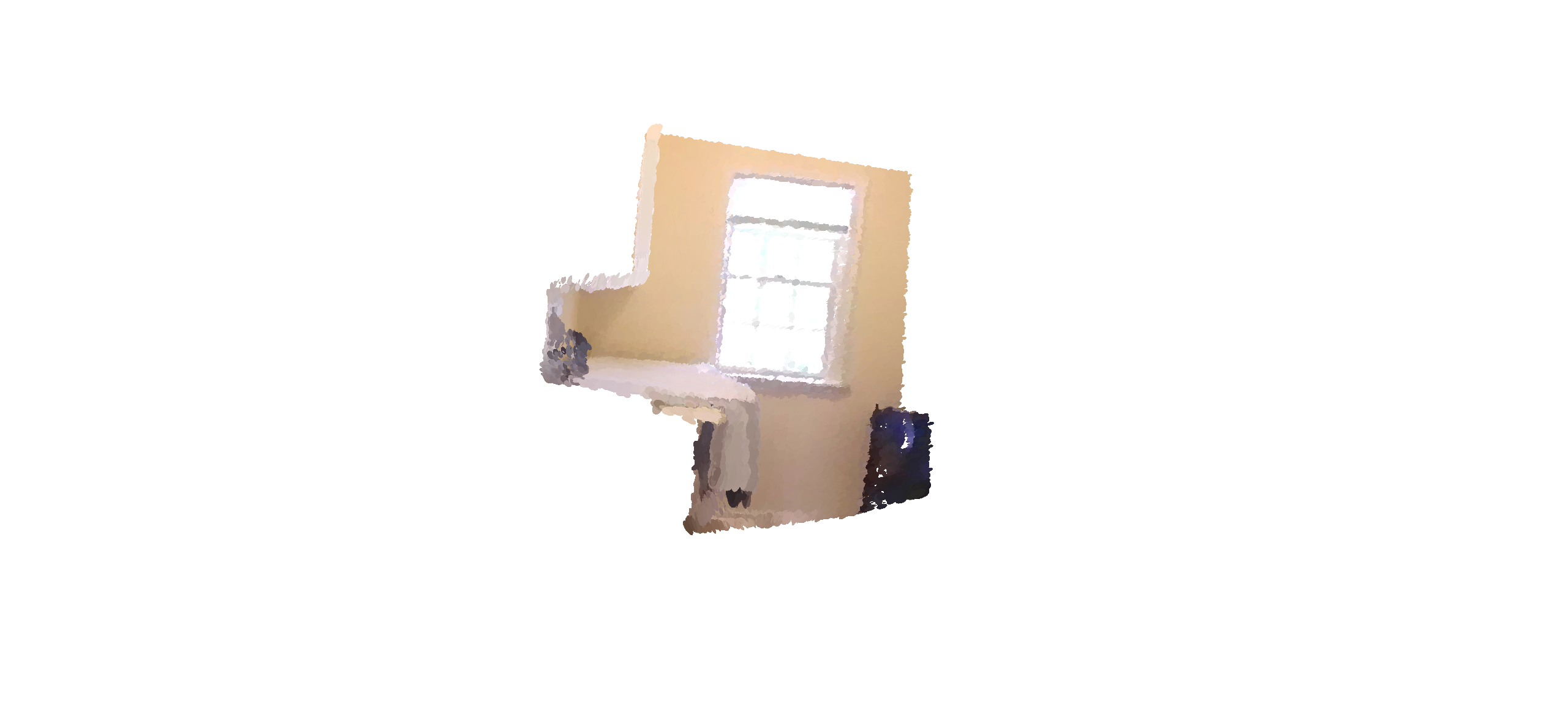}
    \includegraphics[width=0.46\linewidth, clip, trim={900 350 900 250}]{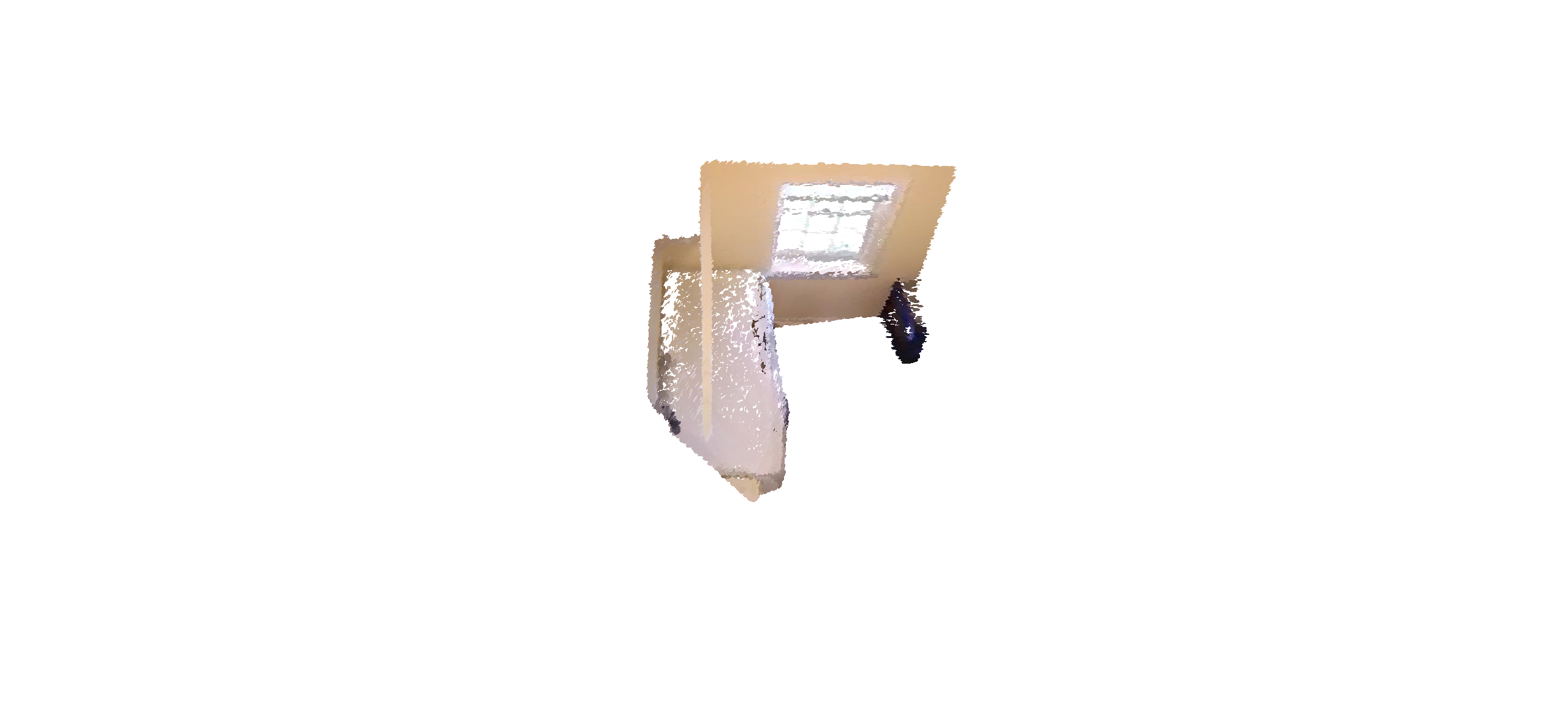}
    \includegraphics[width=0.46\linewidth, clip, trim={900 350 900 250}]{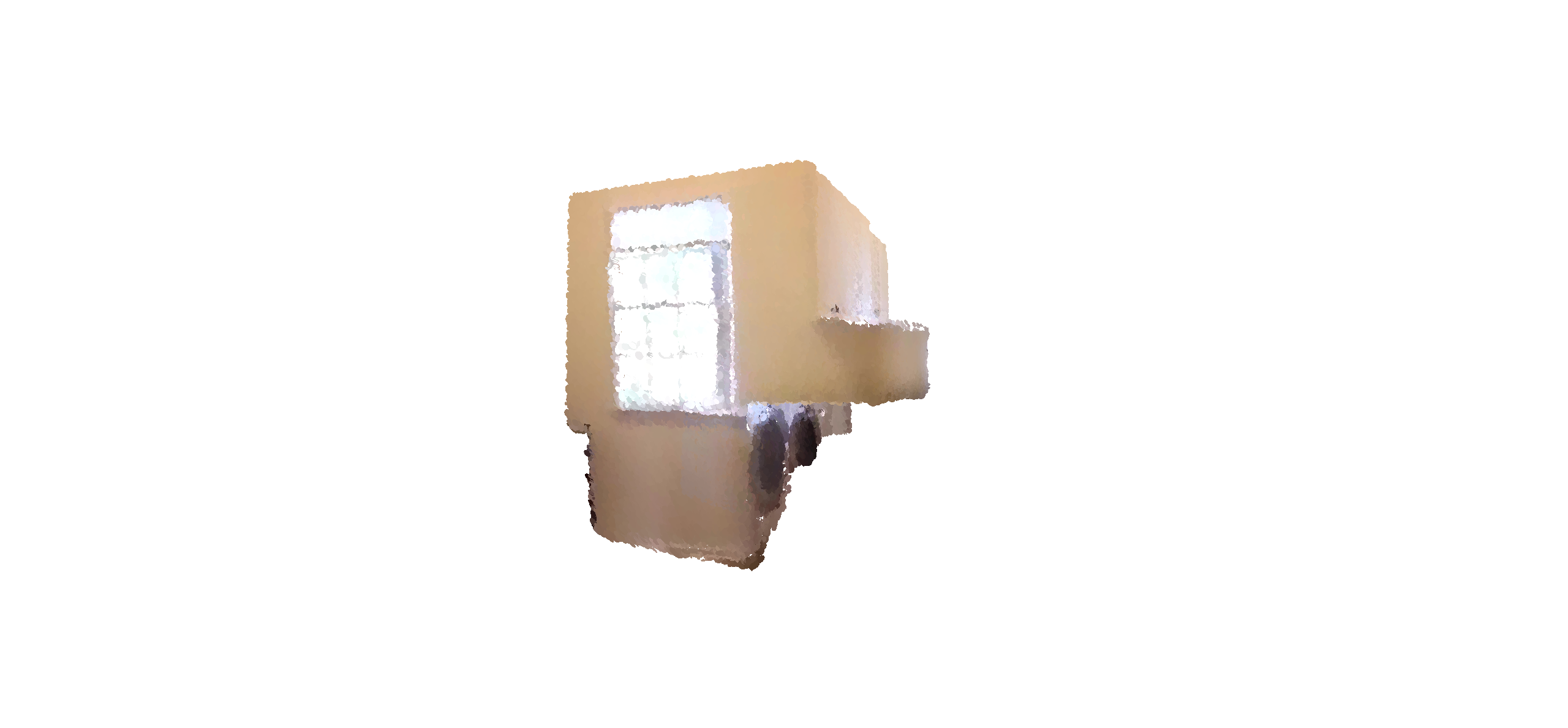}
    \includegraphics[width=0.46\linewidth, clip, trim={900 350 900 250}]{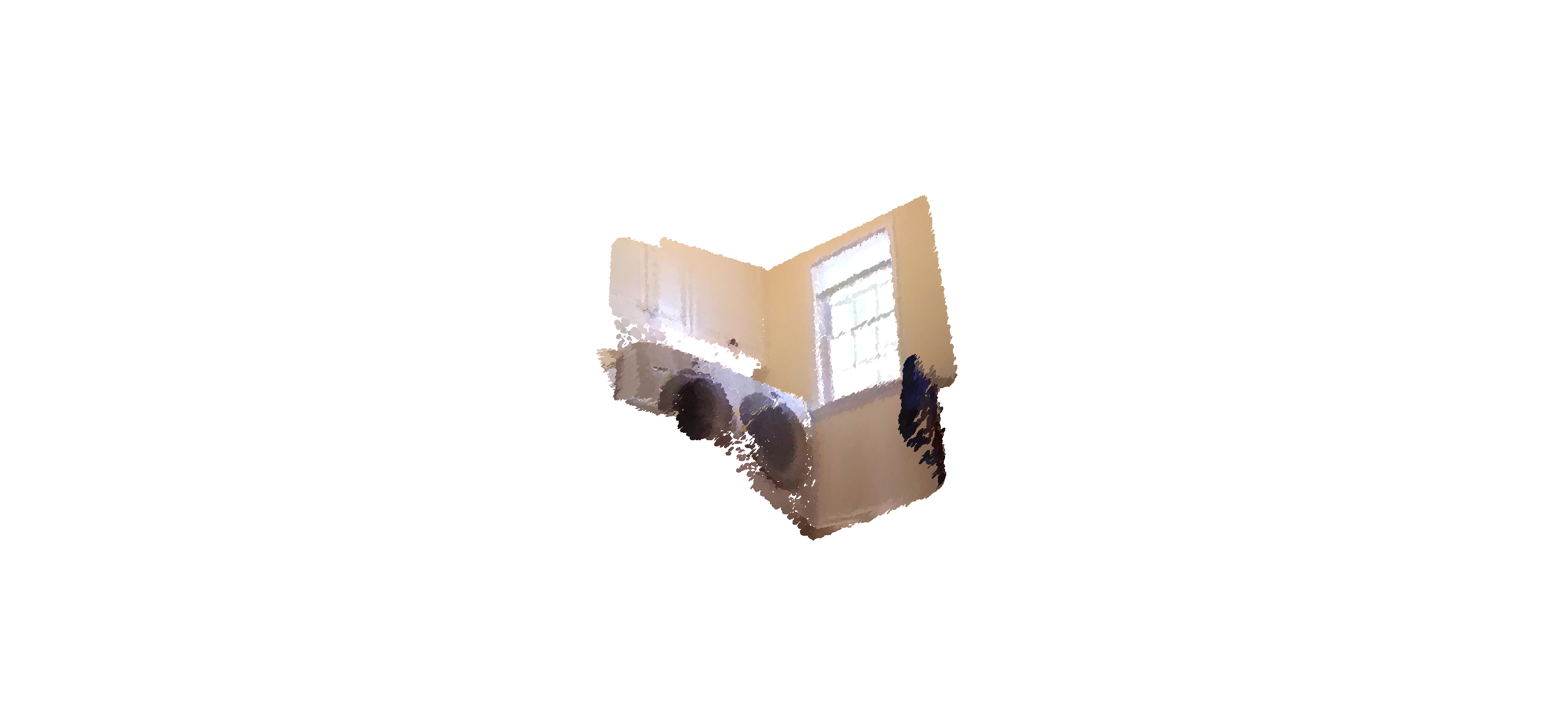}
    \subcaption{Multi-view visualization.}
    \end{subfigure}
    \caption{Visualization for the optimization process and Gaussian Splatting fields.}
    \label{fig:3dscene}
\end{figure*}

\begin{figure*}
\centering
\begin{tikzpicture}
        \node[] at (0, 0) {};
        \node[] at (-7.5, 0) {};
        \node[] at (-6.5, 0) {View 1};
        \node[] at (-6.5+1.5, 0) {View 2};
        \node[] at (-6.5+1.5*2+0.2, 0) {View 1};
        \node[] at (-6.5+1.5*3+0.2, 0) {View 2};
        \node[] at (-6.5+1.5*4+0.4, 0) {View 1};
        \node[] at (-6.5+1.5*5+0.4, 0) {View 2};
        \node[] at (-6.5+1.5*6+0.6, 0) {View 1};
        \node[] at (-6.5+1.5*7+0.6, 0) {View 2};
        \node[] at (-6.5+1.5*8+0.8, 0) {View 1};
        \node[] at (8, 0) {View 2};
    \end{tikzpicture}\\
 \resizebox{0.034\linewidth}{!}{
     	\begin{tikzpicture}
        \node[rotate=270,] at (0, 0) {};
        \node[rotate=270,] at (0, 0.5) {\tiny GT View};
        \node[rotate=270,] at (0, 1.73) {\tiny Rendered View };
        \node[rotate=270,] at (0, -0.7) {\tiny Difference};
        \node[rotate=270,] at (0, -0.8) {};
    	\end{tikzpicture}}
    \includegraphics[width=0.18\linewidth]{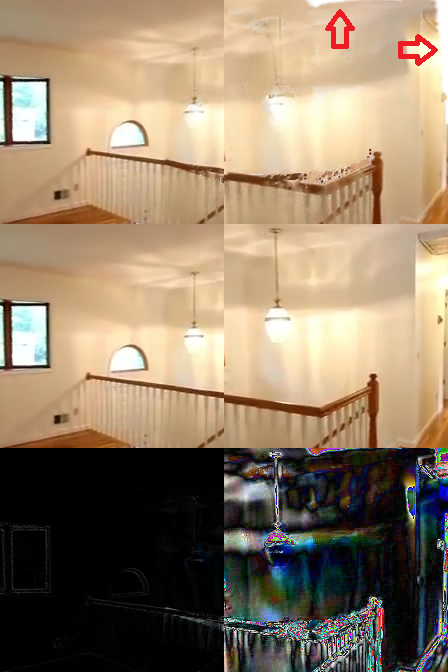}
    \includegraphics[width=0.18\linewidth]{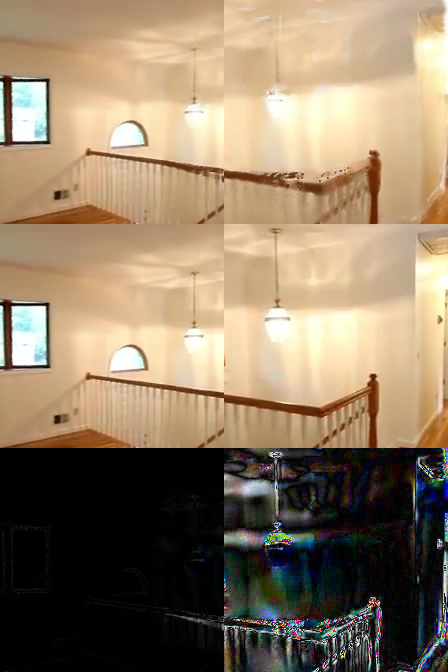}
    \includegraphics[width=0.18\linewidth]{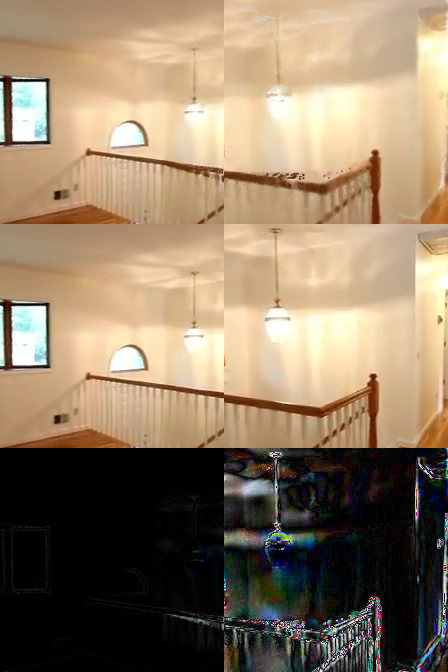}
    \includegraphics[width=0.18\linewidth]{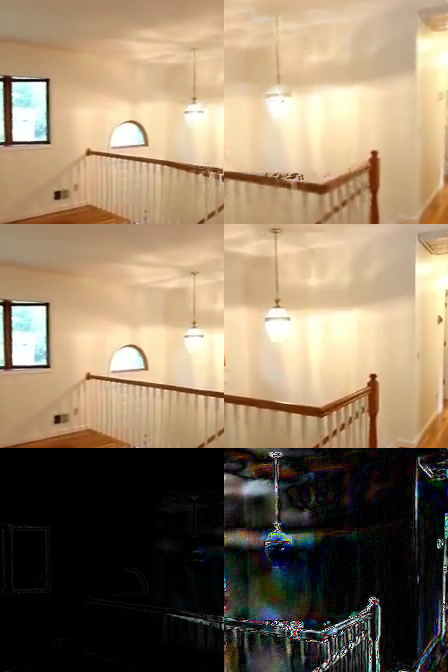}
    \includegraphics[width=0.18\linewidth]{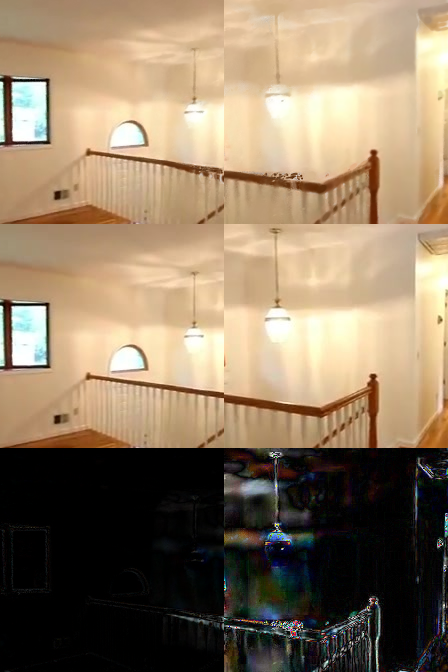}
    \begin{tikzpicture}
        \node[] at (0, 0) {};
        \node[] at (-7.5, 0) {};
        \node[] at (-6.8+1.1, 0) {I=50};
        \node[] at (-6.8+1.1*4-0.1, 0) {I=60};
        \node[] at (-6.8+1.1*7-0.2, 0) {I=70};
        \node[] at (-6.8+1.1*10-0.3, 0) {I=90};
        \node[] at (-6.8+1.1*13-0.4, 0) {I=100};
        \node[] at (8, 0) {};
    \end{tikzpicture}\\
    \caption{Bundle adjusting Gaussian Splatting refinement for View 1 and View 2. For clearly visualizing difference maps, we have magnified the error values by a \textbf{factor of 5}.}
    \label{fig:ba_50_it}
\end{figure*}
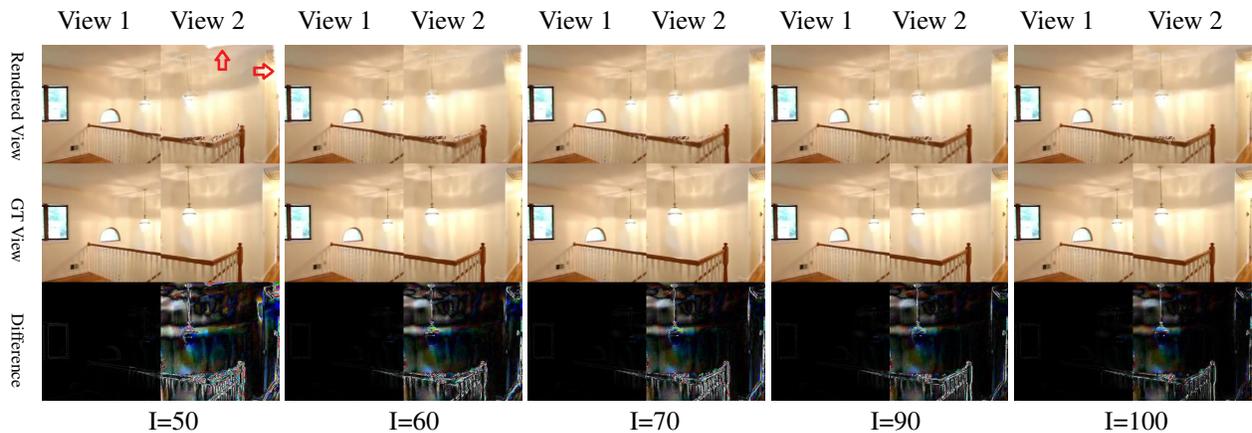

\begin{table}
    \resizebox{\linewidth}{!}{
    \begin{tabular}{cc|cccc|cccc}
    \toprule
       \multirow{2}{*}{Method}  & \ & \multicolumn{4}{c|}{NoPoSlpat~\cite{ye2024no}} & \multicolumn{4}{c}{SmileSplat (Ours)} \\ 
       & \ & Small & Medium & Large & Ave. & Small & Medium & Large & Ave. \\ 
    \midrule
         \multirow{3}{*}{Living 0} & PSNR & 26.80  & 25.01 & 25.71 & 25.84  & \textbf{36.59} & \textbf{35.11} & \textbf{33.21} & \textbf{34.97}\\
         & SSIM & 0.747  & 0.746 & 0.766 & 0.753  & \textbf{0.945} & \textbf{0.971} & \textbf{0.913} & \textbf{0.943} \\
         & LPIPS & 0.298  & 0.276 & 0.256 & 0.277  & \textbf{0.018} & \textbf{0.021} & \textbf{0.029} & \textbf{0.022} \\
    \midrule
         \multirow{3}{*}{Living 1} & PSNR & 27.07  & 25.66 & 25.40 & 26.04  & \textbf{35.44} & \textbf{34.89} & \textbf{34.38} & \textbf{34.90}\\
         & SSIM & 0.701  & 0.748 & 0.761 & 0.736  & \textbf{0.932} & \textbf{0.947} & \textbf{0.943} & \textbf{0.941}\\
         & LPIPS & 0.254  & 0.261 & 0.266 & 0.260  & \textbf{0.020} & \textbf{0.025} & \textbf{0.025} & \textbf{0.023}\\
    \midrule
         \multirow{3}{*}{Office 0} & PSNR & 26.17  & 27.07 & 25.64 & 26.29  & \textbf{36.21} & \textbf{34.10} & \textbf{30.06} & \textbf{33.46} \\
         & SSIM & 0.778  & 0.775 & 0.821 & 0.791  & \textbf{0.973} & \textbf{0.985} & \textbf{0.930} & \textbf{0.963} \\
         & LPIPS & 0.234  & 0.240 & 0.285 & 0.253  & \textbf{0.023} & \textbf{0.012} & \textbf{0.053} & \textbf{0.029}\\
    \midrule
         \multirow{3}{*}{Office 1} & PSNR & 26.65  & 25.21 & 21.30 & 24.39  & \textbf{35.66} & \textbf{34.27} & \textbf{32.89} & \textbf{34.27}\\
         & SSIM & 0.788  & 0.771 & 0.727 & 0.762  & \textbf{0.921} & \textbf{0.977} & \textbf{0.931} & \textbf{0.943}\\
         & LPIPS & 0.259  & 0.249 & 0.291 & 0.266  & \textbf{0.020} & \textbf{0.019} & \textbf{0.037} & \textbf{0.025}\\
    \bottomrule
    \end{tabular}}
    \caption{Cross-Dataset generalization tests on the ICL-NUIM~\cite{handa2014benchmark} dataset.}
    \label{tab:cross-dataset-icl}
\end{table}
\section{Comparison of Novel View Rendering}
\subsection{Qualitative and quantitative Results on the ICL-NUIM and ACID dataset }

As mentioned in Section~\textcolor{red}{4.2}, both qualitative and quantitative results on the ICL-NUIM~\cite{handa2014benchmark} dataset are evaluated for NoPoSplat~\cite{ye2024no} and our method.

As shown in Table~\ref{tab:cross-dataset-icl}, four sequences from living room and office room scenes are selected to evaluate these two approaches. Similar to the cross-dataset evaluation in Table~\textcolor{red}{3}, the inputs are grouped into three categories: Small overlap, Medium overlap, and Large overlap. For the average values, our method achieves a PSNR of over $33.4$, while the best PSNR performance of NoPoSplat is below $26.3$, demonstrating an improvement of up to $26\%$. Furthermore, this improvement is also reflected in two additional metrics.

In Figure~\ref{fig:cross-dataset-ICL}, the reference images (two views) are shown in the left column. The third row (\textit{Office 0}) illustrates the example of small-overlap pairs, while the second row (\textit{Living 1}) shows the example of large-overlap inputs. By comparing the second and third rows, we can observe that our method demonstrates robust performance for both small and large overlap scenes. However, the rendering result of NoPoSplat in the third row contains more outliers, as shown in the difference map. For the Medium-overlap case (first row), alignment issues are noticeable in both the rendered image and the difference map. In contrast, our method, SmileSplat, produces more accurate rendering results, benefiting from the proposed bundle-adjusting Gaussian Splatting module, which refines both the estimated intrinsics and extrinsics.

As shown in Figure~\ref{fig:render_ACID}, many natural scenes from the ACID~\cite{liu2021infinite} dataset are used for testing the performance of the proposed method and state-pf-the-art approaches, MVSplat (CamPara-Required) and NoPoSplat.

As shown in Figure~\ref{fig:render_ACID}, a variety of natural scenes from the ACID dataset are used to evaluate the performance of the proposed method, along with two state-of-the-art approaches: MVSplat (CamPara-Required) and NoPoSplat. These scenes, which cover diverse real-world environments, serve as a comprehensive benchmark for comparing the accuracy and robustness of each method under different conditions, such as varying scene complexity, lighting, and object types. Compared to man-made or indoor scenarios, the scenes in the wild present entirely different challenges. However, the performance of the proposed method in both RGB and depth rendering tasks remains robust and accurate. For example, in the third row, two views are captured in a waterfall environment. The rendered RGB image produced by NoPoSplat shows misalignment issues, whereas our method maintains accurate rendering with better alignment and detail preservation.


\section{Ablation studies}
~\label{sec_more_experiments}

\subsection{Iterative Optimization for Intrinsic and Relative Pose Estimation}

In the refinement stage, we first optimize the intrinsic matrix by \textbf{rendering View 1 for the first 10 iterations}, as shown in Figures~\ref{fig:opti_50_36} and \ref{fig:opti_0_36}. During this stage, only the intrinsic matrix is refined, and the Gaussians are not optimized. As a result, we are still able to detect the photometric distance between the rendered and observed images. The quality of the rendered image improves further over the next 10 iterations of rendering View 1.

Once the optimization of the intrinsic matrix is completed, it is fixed during the second view rendering process. In the first 20 iterations of rendering View 2, neither the intrinsic parameters nor the Gaussians are updated based on the photometric error generated from View 2. As shown in the process for handling View 2, the values in the difference map are large at the beginning due to the noisy initial camera pose. However, the photometric distance progressively decreases as the relative pose estimation converges. When comparing the rendered results at iterations $I=8$ and $I=16$, we observe that the large error regions shift, indicating that it is difficult to eliminate all errors by solely adjusting the camera pose. Therefore, we continue to optimize the Gaussians in subsequent iterations, taking the camera poses into account.

In particular, View 2 detects some newly revealed regions, as shown in Figure~\ref{fig:opti_50_36} (rows of View 2). These regions can be further improved by optimizing the Gaussians over a few more iterations. In these novel view rendering steps, following the approach outlined by methods such as~\cite{ye2024no}, we optimize only the camera poses, without incorporating the observed image in the Gaussian optimization process.

\subsection{Bundle Adjusting Gaussian Splatting based on View 1 and View 2}
As shown in Figures~\ref{fig:opti_50_36}(b), \ref{fig:3dscene}, and \ref{fig:ba_50_it}, the results from the last 50 iterations of bundle adjustment are presented. In particular, Figure~\ref{fig:ba_50_it} (\textcolor{red}{Red Arrow}) shows that at the beginning of the optimization stage, some regions are still misaligned due to the initial inaccuracy of the camera pose. As the optimization progresses, the camera pose is refined incrementally, and we can observe that these regions continue to improve, leading to better rendering results by the $100^{th}$ iteration. 

\subsection{Limitation}
As shown in Figure~\ref{fig:ba_50_it}, it is important to note a limitation of this module, as seen in this figure. For the stair guardrail region, while the rendering quality improves, we cannot achieve perfect results due to the limitations of the initial geometry, which is not accurate for the stair guardrail. Additionally, given that the views from View 1 and View 2 are limited, we are unable to significantly improve the geometry in this region, which is also a future direction for further improvement by leveraging additional multi-view geometric constraints into the system.



{
    \small
    \bibliographystyle{ieeenat_fullname}
    \bibliography{main}
}


\end{document}